%% file: ms.tex
\pdfoutput=1 

\documentclass[
    a4paper,
    twocolumn
]{article}

\usepackage{IEK10} 
\usepackage{natbib}

\usepackage{upgreek}
\usepackage{bbm}
\DeclareMathOperator{\spn}{\mathrm{span}}
\newcommand{\Vb}{\mathbf{b}}

\newcommand{\Vv}{\mathbf{v}}

\newcommand{\Vx}{\mathbf{x}}
\newcommand{\Vy}{\mathbf{y}}
\newcommand{\Vz}{\mathbf{z}}
\newcommand{\Vp}{\mathbf{p}}
\newcommand{\Vq}{\mathbf{q}}
\newcommand{\Vr}{\mathbf{r}}

\newcommand{\Ff}{\bm{f}}

\newcommand{\FN}{\bm{N}}
\newcommand{\Vtheta}{\bm{\uptheta}}

\newcommand{\mA}{\mathbf{A}}

\newcommand{\mJ}{\mathbf{J}}

\def\IEK10{
  Institute of Energy and Climate Research,
  Energy Systems Engineering (IEK-10),
  Forschungszentrum J\"ulich GmbH,
  J\"ulich 52425,
  Germany
}
\def\RWTH{
  RWTH Aachen University
  Aachen 52062,
  Germany
}
\def\JARA{
  JARA-ENERGY,
  J{\"u}lich 52425,
  Germany
}
\def\SVT{
  RWTH Aachen University,
  Process Systems Engineering (AVT.SVT),
  Aachen 52074,
  Germany
}
\def\TUM{
  Department of Informatics, 
  Technical University of Munich, 
  Boltzmannstr. 3, 85748 Garching b. Munich, 
  Germany
}
\def\NUS{
  Department of Mathematics, 
  National University of Singapore, 
  Singapore
}
\def\JHU{
  Departments of Applied Mathematics and Statistics \& Chemical and Biomolecular Engineering, 
  Johns Hopkins University, 
  Baltimore, Maryland 21218, USA
}

\def\R2N2{R2N2}

\newcommand{\mytitle}{A Recursively Recurrent Neural Network (R2N2) Architecture \\ 
for Learning Iterative Algorithms}

\newcommand{\affil}{
  \begin{itemize}[leftmargin=3mm, itemsep=0mm]
    \item[$^a$]\IEK10
    \item[$^b$]\RWTH
    \item[$^c$]\JARA
    \item[$^d$]\SVT
    \item[$^e$]\NUS
    \item[$^f$]\TUM
    \item[$^g$]\JHU
  \end{itemize}
}

\def\firstAuthor{Danimir T. Doncevic}
\newcommand{\myauthor}{\firstAuthor$^{a,b}$%
Alexander Mitsos$^{c,a,d}$
Yue Guo$^{e}$
Qianxiao Li$^{e}$
Felix Dietrich$^{f}$
Manuel Dahmen$^{a,*}$ 
Ioannis G. Kevrekidis$^{g,*}$
}
\newcommand{\correspondingauthor}{M. Dahmen, \IEK10 \newline E-mail: m.dahmen@fz-juelich.de
                                  \newline I. Kevrekidis, \JHU \newline E-mail: yannisk@jhu.edu}
\author{\myauthor}

\usepackage[
  colorlinks,
  linkcolor=blue,
  citecolor=blue,
  urlcolor=blue,
  pdftitle={\mytitle},
  pdfauthor={\firstAuthor}
]{hyperref}
\usepackage[capitalise, nameinlink]{cleveref}
\crefname{table}{Tab.}{Tab.}

\begin{document}

\twocolumn[
\begin{@twocolumnfalse}

  \thispagestyle{firststyle}

  \begin{center}
    \begin{large}
      \textbf{\mytitle}
    \end{large} \\
    \myauthor
  \end{center}

  \vspace{0.5cm}

  \begin{footnotesize}
    \affil
  \end{footnotesize}

  \vspace{0.5cm}
    
  \input{sections_rev/00_abstract.tex}
  \vspace{0.5cm}


  \vspace{0.5cm}

  \vspace*{5mm}

\end{@twocolumnfalse}
]

  \newpage

\input{sections_rev/01_Intro}

\input{sections_rev/02_Problem}

\input{sections_rev/03_Method}

\input{sections_rev/04_Results_Lin}

\input{sections_rev/04_Results_NK}
  \input{sections_rev/04_Results_RK}

\input{sections_rev/05_Conclusion}

  \input{sections_rev/06_Outlook}

\input{sections_rev/acknowledgement}
  
  \bibliographystyle{apalike}
  \renewcommand{\refname}{Bibliography}  
  \bibliography{references.bib}

\end{document}


\twocolumn[
\begin{@twocolumnfalse}
  \thispagestyle{firststyle}

  \begin{center}
    \begin{large}
      \textbf{\mytitle}
    \end{large} \\
    \myauthor
  \end{center}

  \vspace{0.5cm}

  \begin{footnotesize}
    \affil
  \end{footnotesize}

  \vspace{0.5cm}

\end{@twocolumnfalse}
]

\newpage

\input{sections_rev/Appendix_A}

\input{sections_rev/Appendix_B}

\input{sections_rev/Appendix_C}

\input{sections_rev/Appendix_D}

  \bibliographystyle{apalike}
  \renewcommand{\refname}{Bibliography}  
  \bibliography{references.bib}

%% file: sections_rev/00_abstract.tex
\begin{abstract}
  Meta-learning of numerical algorithms for a given task consists of the data-driven identification and adaptation of an algorithmic structure and the associated hyperparameters.
  To limit the complexity of the meta-learning problem, neural architectures with a certain inductive bias towards favorable algorithmic structures can, and should, be used.
  We generalize our previously introduced Runge--Kutta neural network to a recursively recurrent neural network (\R2N2) superstructure for the design of customized iterative algorithms.
  In contrast to off-the-shelf deep learning approaches, it features a distinct division into modules for generation of information and for the subsequent assembly of this information towards a solution. 
  Local information in the form of a subspace is generated by subordinate, \textit{inner}, iterations of recurrent function evaluations starting at the current \textit{outer} iterate.
  The update to the next outer iterate is computed as a linear combination of these evaluations, reducing the residual in this space, and constitutes the output of the network.
  We demonstrate that regular training of the weight parameters inside the proposed superstructure on input/output data of various computational problem classes yields iterations similar to Krylov solvers for linear equation systems, Newton-Krylov solvers for nonlinear equation systems, and Runge--Kutta integrators for ordinary differential equations.
  Due to its modularity, the superstructure can be readily extended with functionalities needed to represent more general classes of iterative algorithms traditionally based on Taylor series expansions.
\end{abstract}

%% file: sections_rev/01_Intro.tex
\section{Introduction}\label{sec:intro}

\begin{figure*}[t]
    \centering
    \subcaptionbox{When the architecture and integrator weights are hard-wired (black connections and their weights), a model of the unknown system (thick red squares) can be approximated, e.g., by multi-layer perceptrons.\label{fig:RK_SysIden}}
        {\includegraphics[scale=0.25,width=0.49\textwidth]{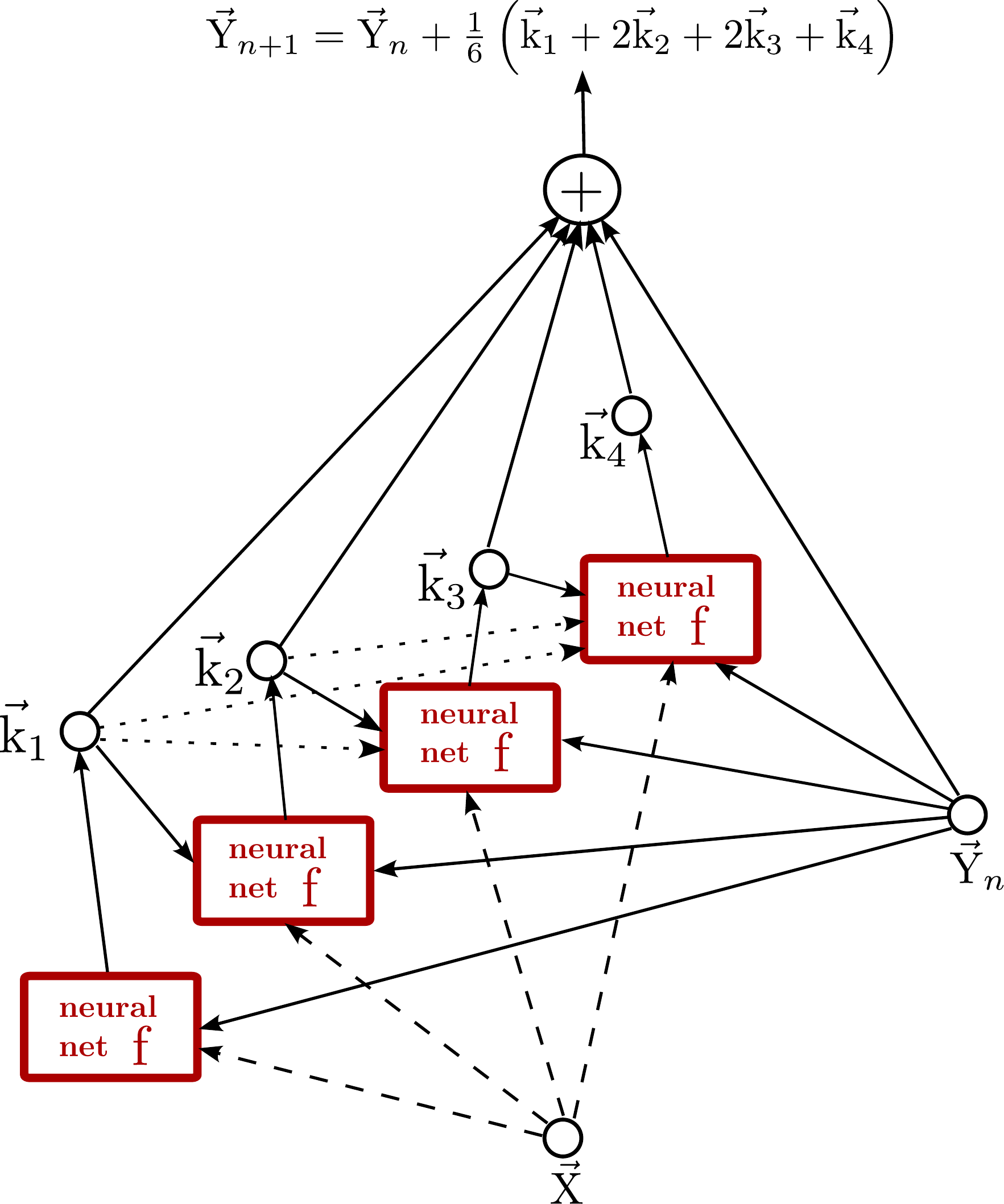}}
    \hfill
    \subcaptionbox{When the model equations are known (black squares), the structure and weights (thick red connections and their weights) associated with a solution algorithm can be discovered for that system.\label{fig:RK_template_Algo}}
        {\includegraphics[scale=0.25,width=0.49\textwidth]{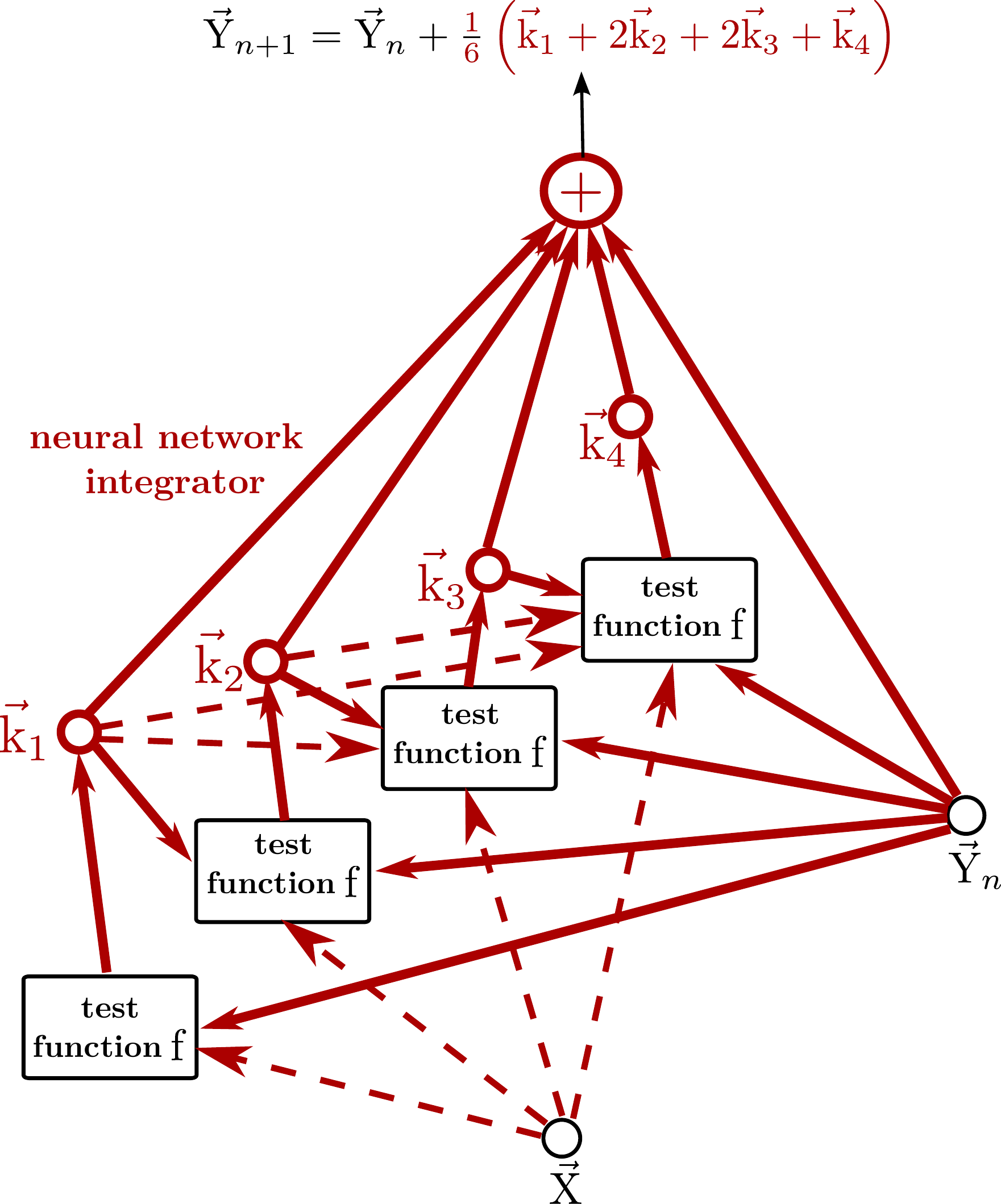}}
    \caption{A recurrent neural network architecture templated on Runge--Kutta integrators, adapted from \cite{RicoMartinez1992discretecontinuous}.
    The illustration shows the classical fourth-order Runge--Kutta scheme, where $\vec{\mathrm{Y}}$ and $\vec{\mathrm{X}}$ are the dependent and independent variables, respectively, $\mathrm{f}$ is the RHS of an ODE, and $\vec{\mathrm{k}}_i$ are stage values.
    Recurrent function evaluations are computed internally, and the network itself can be applied iteratively via external recurrence.
    Given input and output data for training, a model (a) or an algorithm (here, an initial value solver) for that model (b) can be learned.}
    \label{fig:RK_algo_and_SysIden}
\end{figure*}

The relationship between a class of residual recurrent neural networks (RNN) and numerical integrators has been known since \cite{RicoMartinez1992discretecontinuous} proposed the architecture shown in \Cref{fig:RK_SysIden} for nonlinear system identification. 
Together with more recent work of the authors \citep{Mitsos2018algorithms,Guoyue2021metalearning} it motivates the present work.
The salient point of the RNN proposed by \cite{RicoMartinez1992discretecontinuous} is that it provides the structure of an integrator, in that case a fourth-order explicit Runge--Kutta (RK) scheme, within which the right-hand-side (RHS) of an ordinary differential equation (ODE) can be approximately learned from time series data.
To this end, the integrator is essentially hard-wired in the forward pass of the RNN -- a first instance of the body of work nowadays referred to as ``neural ordinary differential equations'' \citep{Chen2018neural}.
However, upon closer inspection, the architecture reveals a complementary utility: When the model equations are known, it is the \textit{structure and weights associated with an algorithm} that can be discovered, see \Cref{fig:RK_template_Algo}.
We study the latter setting in this work.
Algorithms encoded by this architecture inherit two structural features from a RK scheme: i) they are based on recurrent function evaluations (inner recurrence), and ii) they compute iterates by adding a weighted sum of these inner evaluations that acts as \textit{correction} to the input (outer recurrence).
One critical observation is that such nested recurrent features also characterize (matrix)-free Krylov subspace methods such as restarted GMRES \citep{Saad1986GMRES} or Newton--Krylov--GMRES \citep{Kelley1995iterative}, where finite differences (i.e., linear combinations) between recurrent function evaluations provide estimates of directional derivatives in various directions.
This estimation of derivatives, or, more broadly, of Taylor series as in the RK case, forms the backbone of either of these two classical algorithm families, RK and Krylov methods. Observing these parallels, we conjecture that a recursively recurrent neural network (\R2N2) can approximate many traditional numerical algorithms based on Taylor series expansions. 

Returning to \Cref{fig:RK_template_Algo}, we see that the \mbox{(hyper-)}parameters of the \R2N2 are naturally linked to those of the algorithm it furnishes, e.g., weights connecting to the output can encode quadratures, and the internal recurrence count determines the number of function evaluations and the dimension of the subspace that is generated.
It is this particular connectivity, highlighted in red, that is subject to discovery, whether it be encoding a Butcher tableau in the case of RK \citep{Butcher.2016,Guoyue2021metalearning} or finding the combination of weights that convert function evaluations into directional derivatives for \mbox{Newton--Krylov} (NK).

The question that motivates this work is thus:
\textit{Can meta-learning applied to the architecture and parameters of the \R2N2 discover old and new algorithms ``personalized'' to certain problem classes?}
We advocate that the \R2N2 provides an architecture search space particularly suited for answering this question. 
The parameters of the \R2N2 can be partitioned \textit{a priori} into those that are hard-wired and those subject to meta-learning. We can focus on the discrete features, i.e., determining the operations and connections that yield a best performant algorithm, or on the (continuous) weights, i.e., on how to best combine these operations. The former is also referred to as \textit{algorithm configuration}, while the latter is a special case of \textit{parameter tuning} \citep{Hoos2011automated}.
Algorithm configuration has been optimized in a previous effort of the authors \citep{Mitsos2018algorithms}, albeit for a different, less expressive architecture with all weight parameters collapsed into a scalar stepsize. 
More recently, we worked with a fixed architecture as in \Cref{fig:RK_template_Algo} such that Butcher tableaus personalized to specific classes of initial-value problems (IVPs) were learned \citep{Guoyue2021metalearning}.
The present work seeks to extend this latter application to further numerical algorithms, in particular NK, and to thereby demonstrate that the \R2N2 gives rise to a potent \textit{superstructure} for optimal algorithm discovery.
In follow-up work, we aim to show that jointly optimizing the weights and architecture from such a superstructure realizes learning of optimal iterative numerical algorithms, and to explore the use of approximate physical models as preconditioners within the superstructure.

\subsection{Related work}

Automated algorithm configuration and tuning dates back to \cite{Rice1976algorithm} and studies meta-algorithms or meta-heuristics that pick the best algorithm from a set of (parametrized) algorithms for certain problem classes by manipulating the (hyper)parameters of a solver \citep{Hoos2011automated}. 
Speed-ups up to a factor of $50$ have been demonstrated, e.g., for satisfiability problems \citep{Khudabukhsh2016satenstein} or mixed-integer problems \citep{Hutter2010automated} by leveraging problem structure. 
This is in contrast to classical numerical algorithms that are designed to work with little or no \textit{a priori} knowledge about the internal structure of the problem instances to be solved, and are often biased towards worst-case performance criteria \citep{Gupta2020datadriven}. In contrast, algorithms adapted to specific problem classes typically incur a generalization weakness on other problems \citep{Wolpert1997nofreelunch}.
\cite{Mitsos2018algorithms} modeled algorithms as feedback schemes with a cost ascribed to each operation, such that the design of iterative algorithms was posed as an optimal control problem of mixed-integer nonlinear (MINLP) type. They restricted operations to monomials of function evaluations and derivatives, thus specifying a family of algorithms. Depending on the analyzed problem, the procedure would recover known, established algorithms from this family, but also new algorithms that were optimal for the problems considered. 

\cite{Mitsos2018algorithms} solved the algorithm generation MINLPs via the deterministic global solver BARON \citep{Tawarmalani2005polyhedral}.
Possible gains of tailored algorithms must be weighed up against the required effort for finding them. Machine learning (ML) can decrease this effort.
This has recently been demonstrated forcefully by \cite{Fawzi2022discovering} who improved the best known algorithms on several instances of matrix multiplication, an NP-hard problem, and \cite{Mankowitz2023faster} who found new state-of-the-art sorting algorithms, both using deep reinforcement learning built on top of the AlphaZero framework \citep{Silver2018general}.
This landmark achievement is expected to spur new interest in algorithm discovery via ML.
ML methods can optimize the performance of algorithms on a problem class implicitly given through data \citep{Balcan2020datadriven, Gupta2020datadriven}.
Such data-driven algorithm design has been applied to several computational problems, e.g., learning to solve graph-related problems (e.g., \citep{Tang2020towards}), learning sorting algorithms (e.g., \citep{Schwarzschild2021can}), learning to branch \citep{Khalil2016learning,Balcan2018learning}, meta-learning optimizers (e.g., \citep{Andrychowicz2016learning, Metz2020tasks}), and our previous work of meta-learning RK integrators \citep{Guoyue2021metalearning}.
In the context of this problem, the \R2N2 defines a \textit{superstructure} for a class of iterative algorithms.
The notion of superstructure is used in many disciplines to denote a union of structures that are candidate solutions to a problem, e.g., in optimization-based flowsheet design within process systems engineering \citep{Yeomans1999systematic,Mencarelli2020superstructure}.
In general, optimizing a superstructure requires \textit{integer} optimization techniques as in \cite{Mitsos2018algorithms}. Given the neural network interpretation of the \R2N2, however, (heuristic) methods for neural architecture search, e.g., \citep{Elsken2018efficient,Li2020geometry}, may be considered. Consequently, the optimal algorithmic procedure corresponds to an optimized neural architecture. 
In this work, we avoid integer optimization by essentially fixing the neural architecture for each respective numerical experiment, and optimizing the weights therein. 

The \R2N2 belongs to the class of recurrent neural networks (RNNs), which are a natural fit for iterative algorithms.
For instance, RNN architectures templated on RK integrators have been suggested several decades ago \citep{RicoMartinez1992discretecontinuous,RicoMartinez1994continuoustime, RicoMartinez1995nonlinear,GonzalezGarcia.1998}. 
These architectures can be used for both, computing the outputs of an integrator, e.g., \citep{RicoMartinez1992discretecontinuous, GonzalezGarcia.1998} and identifying terms in a differential equation, e.g., \citep{RicoMartinez1994continuoustime,Nascimento2020tutorial, Lovelett2020partial, Zhao2020discovery, Goyal.2021}. 
Further, as the RK network of \cite{RicoMartinez1992discretecontinuous} learns the residual between the input and the output data, it constitutes the first occurrence of a \textit{residual} network (ResNet, \cite{He2016resnet}).
Several newer ResNet-based architectures retain structural similarities with numerical methods \citep{Lu2018beyond}. 
For instance, FractalNet \citep{Larsson2016FractalNet} resembles higher-order RK schemes in its macrostructure, as does the ResNet-derived architecture by \cite{Schwarzschild2021can}.
\cite{Lu2018beyond} proposed a ResNet augmented by a module derived from linear multistep methods \citep{Butcher.2016}. 
Beyond architectural similarity with numerical methods, neural networks proposed in literature can also be designed such that the mathematical mapping they represent shares desirable properties with that of a method, e.g., symplectic mappings \citep{Jin2020sympnets} for symplectic integrators or contracting layers \citep{Chevalier2021contracting} for fixed-point iterations.
\cite{Dufera2021deep} and \cite{Guoyue2021metalearning} trained networks to match the derivatives of ODEs or of their RK-based solution expansion at training points, respectively.
Finally, direct learning of algorithms from neural network-like graphs has been proposed, e.g., by \cite{Tsitouras2002neural}, \cite{Denevi2018learning}, \cite{Mishra2018machine} and \cite{Venkataraman2021neural}. 

\subsection*{Contributions}

We build on the \mbox{RK-NN} of our previous work \citep{Guoyue2021metalearning} and introduce the \R2N2 superstructure for iterative numerical algorithms. 
The function to be evaluated inside the \R2N2 architecture is itself explicitly given as part of the input problem instance, which is a major difference to operator networks that learn parameter-to-solution mappings like those in \cite{Lu.2019} and \cite{Li2020fourier}. Thus, higher-order iterative algorithms for equation solving and numerical integration, that are traditionally constructed through Taylor series expansion, can be approximated by the \R2N2 superstructure.
We demonstrate that both NK methods for solving systems of equations and RK methods for solving IVPs are encompassed by the proposed \R2N2 as a joint superstructure.
Further, in numerical experiments we show that the trained \R2N2 can match, and sometimes improve upon, the iterations performed by NK and RK algorithms for a given number of function evaluations. A comparison of the \R2N2 to GMRES on linear equation systems -- which are the basic building block of iterative solvers for nonlinear systems -- provides insight into the operations of the \R2N2.
In these experiments, our particular realization of this superstructure has a strong inductive bias that alleviates the need for certain configuration decisions, i.e., integer optimization, in algorithm design. The weights that remain to be trained correspond to coefficients or hyperparameters of the algorithms in question, and we tune these using PyTorch \citep{Paszke2019pytorch}. 

The remainder of this article is structured as follows.
In \Cref{sec:preliminaries}, we give a general problem definition for learning algorithms from task data and specify the problem for iterative algorithms.
\Cref{sec:method} introduces the \R2N2 superstructure for learning iterative algorithms and shows its relation to steps of iterative equation solvers and integrators.
\Cref{sec:results} presents results of numerical experiments, where the \R2N2 is trained to perform iterations of linear and nonlinear equation solvers and integrators. 
We summarize our results in \Cref{sec:conc} and discuss future research opportunities in \Cref{sec:outlook}.

%% file: sections_rev/02_Problem.tex
\section{Problem definition}\label{sec:preliminaries}

Various generic problem formulations for learning algorithms from problem data exist in the literature, e.g., in \cite{Balcan2020datadriven}, \cite{Gupta2020datadriven}, and our prior work \citep{Guoyue2021metalearning}.
This section provides background and notation for a generic algorithm learning problem, \Cref{sec:pproblem}, and specifies the problem formulation for learning iterative algorithms, \Cref{sec:piterative}.

\subsection{Generic problem formulation}\label{sec:pproblem} 

Let $\Vx \in \mathbb{R}^{m}$ and $\mathcal{F}$ a set of vectors such that $F \in \mathcal{F}$ is a problem instance composed of a continuous function $\Ff \vcentcolon \mathbb{R}^{m} \mapsto \mathbb{R}^{m}$ and some optional, additional problem parameters $\Vp$, e.g., time, such that we can write $F=\left(\Ff, \Vp \right)$.
Then, a traditional class of problems can be characterized by a functional $\Pi$ acting on $F$ and a point in $\mathbb{R}^m$. 
Further, a \textit{solution} of $F$ is any $\Vx^{\star}\in \mathbb{R}^m$ for which
\begin{equation}\label{eqn:generic_task}
    \Pi(\Vx^{\star}, F) = 0.
\end{equation}
This abstract form encompasses many problem types. The first type we consider here is finding the solution $\Vx^{\star}$ of algebraic equations, s.t
\begin{equation}\label{eqn:algebraic_fp}
    \Pi(\Vx^{\star},F) = \left\|\Ff(\Vx^{\star})\right\| = 0.
\end{equation}
The second problem type is finding the solution of initial-value problems (IVPs) for a specific end time $t$, s.t.
\begin{equation}\label{eqn:integration_fp}
    \Pi(\Vx^{\star},F) = \left\| \Vy_0 + \int_{\tau=t_0}^{\tau=t}\Ff(\Vy(\tau))d\tau - \Vx^{\star} \right\|= 0, 
\end{equation} 
where $\Ff$ is the RHS of an ODE and $\Vy(\tau) \in\mathbb{R}^m$ is the state of the system at time $\tau$ with the initial value $\Vy_0 =\Vy(\tau=t_0)$. 
Equation~\eqref{eqn:integration_fp} illustrates the definition of a problem instance and optional parameters $\Vp$, where $\Vp =\left(\Vy_0,t\right)$. The solution is the final value of the states, i.e., $\Vx^{\star}=\Vy(\tau=t)$.

We only consider problem instances $F$ for which a solution $\Vx^{\star}$ exists.
Then, we call any operator mapping $F \mapsto \Vx^{\star}$ a solver $\mathcal{S}(F)$. 
In general, we do not explicitly know these solvers and hence need to approximate their action by the design and use of numerical algorithms. 
That is, algorithms act as approximate solvers $\mathcal{A}(F, \Vtheta) \approx \mathcal{S}(F)$, parametrized by $\Vtheta$, such that
\begin{equation*}
    \Pi\left(\mathcal{A}(F,\Vtheta),F \right) \leq\delta,
\end{equation*} 
where $\delta \in \mathbb{R}^{+}$ is sufficiently small and ideally user-defined.
The challenge of designing numerical methods pertains to finding a structure linking mathematical operations parametrized by a set of parameters $\Vtheta$, which together yield performant solvers for problems that can be recast to Problem~\eqref{eqn:generic_task}. 

Due to the effort expended in designing algorithms, another important consideration is the range of applicability of the algorithm, i.e., the size of the set of problems it can solve.
Traditional algorithms for problems like \eqref{eqn:algebraic_fp} and \eqref{eqn:integration_fp} often consider a certain worst-case performance in a given problem class $\mathcal{F}$.
In contrast, we are interested in the average/expected performance of algorithms over a specific problem class, which is defined by a distribution $\mu$ over $\mathcal{F}$.
Then, finding performant, or even optimal, algorithms for such a class of problems requires solving
\begin{equation}\label{eqn:generic_learningproblem}
    \min_{\Vtheta} \mathbb{E}_\mu \left[ \left\Vert \Pi\left( \mathcal{A}(F,\Vtheta), F \right) \right\Vert \right] +R(\mathcal{A}(F,\Vtheta),\Vtheta).
\end{equation}
Problem~\eqref{eqn:generic_learningproblem} seeks parameters $\Vtheta$ which minimize the expectation of the residual norm of solutions computed to Equation~\eqref{eqn:generic_task} using $\mathcal{A}(F,\Vtheta)$ over problem instances distributed according to $\mu$. 
The second term in \eqref{eqn:generic_learningproblem}, $R(\mathcal{A}(F,\Vtheta),\Vtheta)$, is reserved for some additional regularization penalty that can promote certain properties in the approximate solution $\mathcal{A}(\cdot,\Vtheta^{\star})$, such as a desired convergence order \citep{Guoyue2021metalearning}.
The regularization can also penalize the parameter values $\Vtheta$ directly, e.g., in $L_2$ regularization. The algorithmic structure itself can be described by discrete variables, e.g., to indicate whether an operation or module exists in the algorithm. 
However, to include such discrete variables one requires a metric that determines which structure is optimal. This is typically assessed over a prolonged number of iterations and requires integer optimization techniques, see, e.g., \cite{Mitsos2018algorithms}.
In this work, the goal is rather to demonstrate that the iterations of several iterative algorithms have a common superstructure.
Consequently, we focus on tuning the parameters of this superstructure towards different problem classes. 
Thus, Problem~\eqref{eqn:generic_learningproblem} resembles a regular multi-task learning problem in the context of statistical learning \citep{Baxter2000model}, where \textit{tasks} are equated with problem instances $F$. 
To handle such a problem, the expectation term can be approximated by drawing samples of task data from $\mu$ and minimizing some loss function for them.

\subsection{Iterative numerical algorithms}\label{sec:piterative}

We focus on iterative algorithms, which construct a sequence of iterates $\{\Vx_k\}$ approaching a solution of $F$. 
Starting with the initial point $\Vx_0$, the algorithm computes new iterates by applying
\begin{equation}\label{eqn:iteration}
    \Vx_{k+1}=\mathcal{A}^{iter}(F,\Vtheta;\Vx_k),
\end{equation}
such that the sequence ideally converges to some $\hat{\Vx}$:
\begin{equation*}
    \hat{\Vx} = \lim\limits_{k \rightarrow \infty}{\Vx_k}. 
\end{equation*}
If we have $\Pi(\hat{\Vx},F) = 0$, then $\mathcal{A}^{iter}$ is convergent to the solution of $F$.

In the following, we restrict the possible realizations of $\mathcal{A}^{iter}$ strongly by narrowing our attention to iterative algorithms that apply \textit{additive} step updates in a generalized Krylov-type subspace $\mathcal{K}_n$. 
This subspace is spanned by vectors $\Vv_0, \ldots, \Vv_{n-1}$, i.e.,
\begin{equation}\label{eqn:RKK_ssDef}
    \mathcal{K}_n(\Ff,\Vx) = \spn \{\Vv_0, \Vv_1, \ldots,\Vv_{n-1}\},
\end{equation}
that are generated by recurrent function evaluations at $\Vx_k$, i.e.,
\begin{subequations}\label{eqn:RKK_member}
\begin{equation}\label{eqn:RKK_member_0}
    \Vv_0 \vcentcolon = \Ff\left(\Vx_k\right),
\end{equation}
\begin{equation}\label{eqn:RKK_member_j}
    \Vv_j = \Ff\left(\Vx_k +\sum_{l=0}^{j-1} b_{jl} \Vv_l \right),  \quad j=1,\ldots, n-1,
\end{equation}
\end{subequations}
for some $b_{jl} \in \mathbb{R}$.
The next iterate $\Vx_{k+1}$ is computed by adding a linear combination of this basis to $\Vx_k$, i.e.,
\begin{equation}\label{eqn:RKK_final}
    \Vx_{k+1} = \Vx_k + \sum_{j=0}^{n-1} c_j \Vv_j,
\end{equation}
for some $c_j \in \mathbb{R}$.
Equation~\eqref{eqn:RKK_member_j} describes an \textit{inner} iteration of the algorithm, while Equation~\eqref{eqn:RKK_final} constitutes an \textit{outer} iteration. 
Together, Equations~\eqref{eqn:RKK_member}---\eqref{eqn:RKK_final} define a common \textit{superstructure} for iterative algorithms. 
This structure has parallels to Krylov subspace methods \citep{Saad2003iterative}, and RK methods \citep{Butcher.2016}, respectively, see \Cref{sec:mcoefficient_matching}. 

Considering only iterative algorithms within this superstructure has several advantages compared to alternatives that rely on deep learning with heavily parametri\-zed models.
First, the iterative nature of the superstructure greatly reduces the total amount of parameters needed to learn solution procedures.
Second, Equations~\eqref{eqn:RKK_member} and \eqref{eqn:RKK_final} eliminate most of the functional forms admissible under Equation~\eqref{eqn:iteration}. And third, directly embedding $\Ff$ in the superstructure avoids the need to learn an extra representation of the problem within the solver mapping.
The resulting superstructure resembles an RNN, and automatic differentiation frameworks for the training of neural networks such as PyTorch \citep{Paszke2019pytorch} can be used to determine the remaining free parameters $\Vtheta$.

%% file: sections_rev/03_Method.tex
\section{\R2N2 superstructure for iterative algorithms}\label{sec:method}

We recently proposed the \mbox{RK-NN}, a neural network architecture templated on RK integrators, to \textit{personalize} coefficients of a RK method to a specific problem class \citep{Guoyue2021metalearning}.
In this work, we extend our view of the \mbox{RK-NN} to that of a more general superstructure for iterative numerical algorithms, the \R2N2, that is applicable to different problem classes such as equation solving and numerical integration.
\Cref{sec:msuperstruct} describes the architecture of the \R2N2. Each forward pass through the \R2N2 is interpreted as an iteration of a numerical algorithm (\textit{outer} recurrence) that invokes one or many recurrent function evaluations (\textit{inner} recurrence), starting at the current iterate, to compute the next iterate. The \R2N2 is equivalent to the \mbox{RK-NN} from \cite{Guoyue2021metalearning} for the specific case of minimizing empirical risk (Problem~\eqref{eqn:generic_learningproblem}) for classes of IVPs, Problem~\eqref{eqn:integration_fp}. 
However, as a small addition to the original \mbox{RK-NN}, the \R2N2 superstructure now allows presetting various routines for function evaluation, see supplementary material (\textbf{SM1}). 
In \Cref{sec:mcoefficient_matching}, we show that the applicability of the \R2N2 as a superstructure is substantially extended beyond the case of solving IVPs.
\Cref{sec:mloss} and \Cref{sec:mimplementation} conclude this section with remarks about training and implementation of the \R2N2 superstructure.

\subsection{Neural architecture underlying the \R2N2 superstructure}\label{sec:msuperstruct}

\begin{figure*}[htb] 
    \centering
    \includegraphics[scale = 0.45]{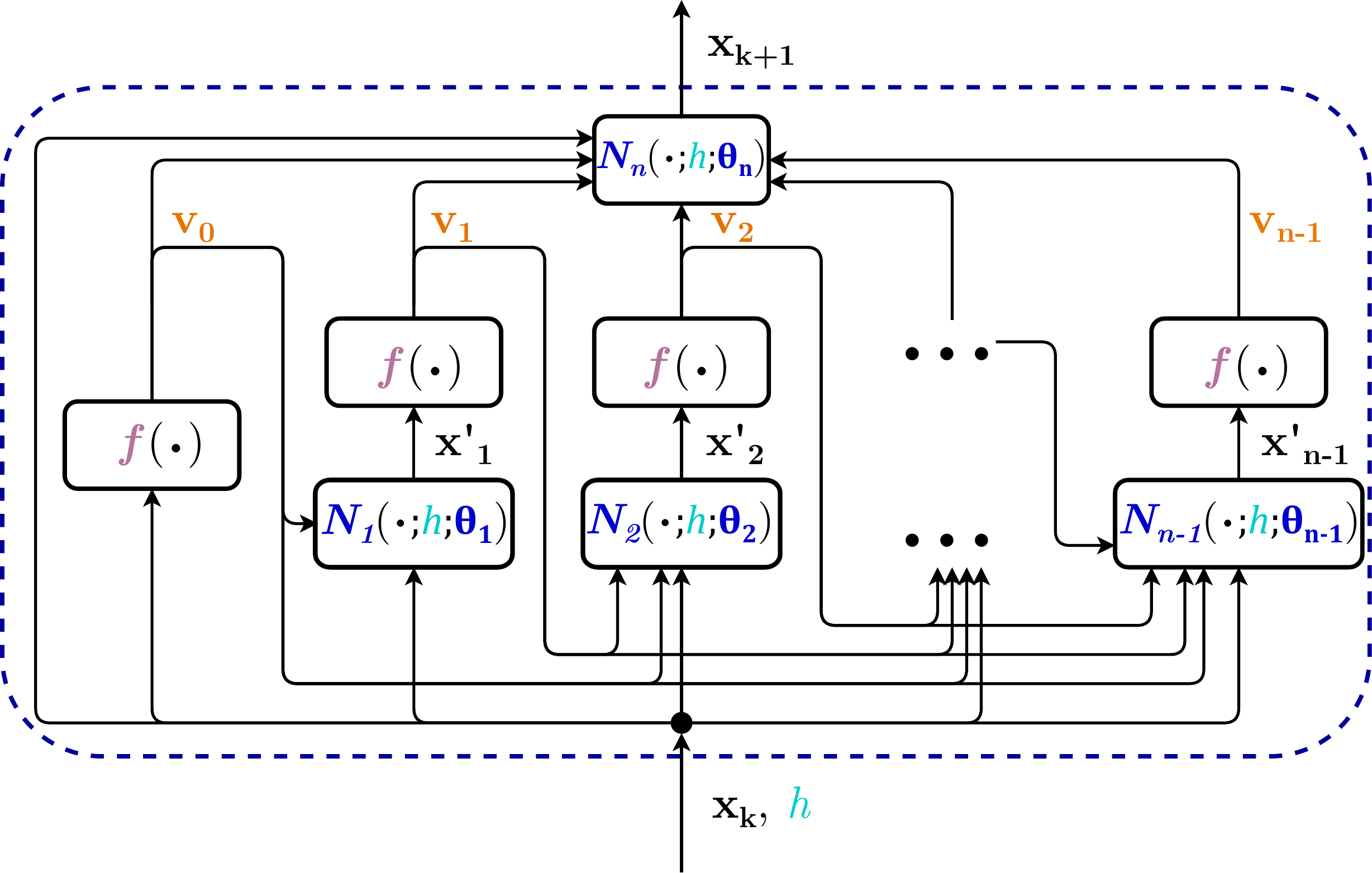}
    \caption{Proposed recursively recurrent neural network-based superstructure of an iterative algorithm. 
    The recurrent cell is delimited by the dashed blue line and computes $\Vx_{k+1}$ as a function of the current iterate $\Vx_k$, a scaling parameter $h$ (cyan), and a function $\Ff$ (magenta). 
    Trainable weights $\Vtheta_j$ are contained within each of the layer modules $\FN_j$ with $j=1,2,\ldots,n-1$. $\Vtheta_n$ denote the trainable weights of the output module $\FN_n$ (all in blue). 
    $\Vx_k$ has a skip connection to each layer and to the output module. 
    Intermediate function arguments, i.e., inputs to $\Ff$ in layer $j$, are called $\Vx'_j$ and subspace members, i.e., outputs of $\Ff$ in layer $j$, are denoted by $\Vv_j$ (orange).}
    \label{fig:superstruct}
\end{figure*}

\begin{figure}[htb] 
    \centering
    \includegraphics[scale = 0.45]{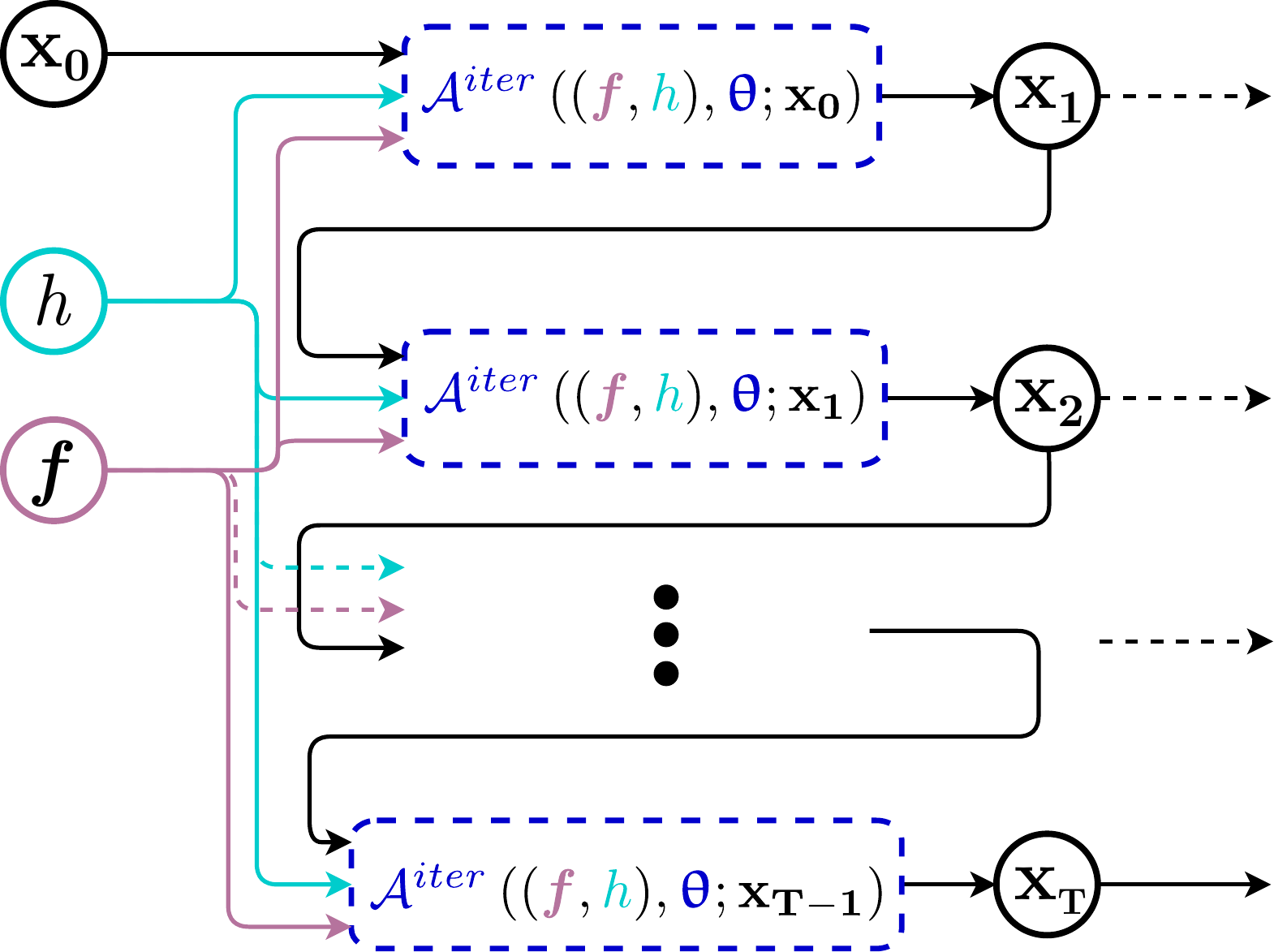}
    \caption{$T$ consecutive iterations using the \R2N2 superstructure from \Cref{fig:superstruct} compute a sequence $\left(\Vx_1, \ldots, \Vx_T \right)$. The blue dashed cell delimits the iteration $\mathcal{A}^{iter}$ computed in a forward pass, i.e., one pass through the \R2N2. The task-related inputs $h$ (cyan) and $\Ff$ (magenta) apply to each iteration. The initial guess $\Vx_0$ is an input only to the first iteration.}
    \label{fig:superstruct_iters}
\end{figure}

The proposed \mbox{\R2N2} superstructure is portrayed by \Cref{fig:superstruct}. It represents the computation of one iteration of a numerical algorithm, i.e., the mathematical function defined by the RNN architecture can substitute for $\mathcal{A}^{iter}(F,\Vtheta;\Vx_k)$ in Equation~\eqref{eqn:iteration}, where a problem instance $F$ contains $\Ff$ and $\Vp$. Each step requires the current iterate $\Vx_k$ as an input, where the initial $\Vx_0$ is typically supplied within $\Vp$. 
Further, a function $\Ff \vcentcolon \mathbb{R}^m \mapsto \mathbb{R}^m$ is prescribed externally as part of a task, but remains unchanged for all iterates. 
Finally, a parameter $h$ for scaling of the layer computations is part of the input to the superstructure.
For some problem classes, we choose $h$ according to problem parameters $\Vp$, e.g., the timestep in integration. Whenever $h$ is not specified, assume $h=1$, i.e., no scaling.
The initial layer, which is the left-most in \Cref{fig:superstruct}, is always a direct function evaluation at the current iterate, $\Ff(\Vx_k)$, i.e., we have $\Vv_0 = \Ff(\Vx_k)$.
The remaining $n-1$ layers of the superstructure each output a $\Vv_j$, $j=1,\ldots, n-1$ by applying a composition of $\Ff$ and $\FN_j$.
$\FN_j$ linearly combines its inputs $\Vv_0,\ldots,\Vv_{j-1}$ using trainable parameters $\Vtheta_j$ and, scales the term with $h$ and adds it to $\Vx_k$ to provide $\Vx'_j$, the input to $\Ff$ in the $j$-th layer:
\begin{equation}\label{eqn:RKNN_module}
    \Vx'_j =\FN_j \left(\Vx_k;\Vv_0,\ldots,\Vv_{j-1}; h \right) = \Vx_k+ h\sum_{l=0}^{j-1} \theta_{j,l} \Vv_l 
\end{equation}
The $\{\Vv_j\}$, including $\Vv_0$, span an $n$-dimensional subspace in which the output layer $\FN_n$ computes the next iterate $\Vx_{k+1}$ using the trainable parameters $\Vtheta_n$, i.e.,
\begin{equation}\label{eqn:RKNN_final}
    \Vx_{k+1} = \Vx_k + h \sum_{j=0}^{n-1} \theta_{n,j} \Vv_j.
\end{equation}
This completes one forward pass through the \R2N2 superstructure.
Note that parameters $\Vtheta$ are partitioned into $\left\{ \Vtheta_1,\ldots,\Vtheta_{n-1}, \Vtheta_n\right\}$ and that we can identify $h \theta_{j,l}$ with $b_{jl}$ from Equation~\eqref{eqn:RKK_member_j} for $j = 1, \ldots, n-1$ 
and $\theta_{n,j}$ with $c_j$ from Equation~\eqref{eqn:RKK_final}.
\Cref{fig:superstruct_iters} shows the computation of multiple consecutive iterations with the proposed \R2N2 superstructure.

\subsection{Relation to Krylov subspace methods}\label{sec:mcoefficient_matching} 

Numerical algorithms for solving (non-)linear equations as well as integrating sets of ordinary differential equations are traditionally based on local Taylor series expansions, and on the use of the first term (the Jacobian) -- or even sometimes the second term (the Hessian) -- in performing the numerical operations required to arrive at the next iteration. 
The solution of sets of linear equations resulting from sets of nonlinear equations, e.g., Newton iterations involving the solutions of linear equations, underpins a lot of today's scientific computing. 
Due to the memory constraints of the early supercomputers (like the Cray 1), algorithms solving sets of linear equations through matrix-vector products and the construction of Krylov subspaces, e.g., GMRES, blossomed in the 1970s and 80s. 
Most importantly for us, these ideas evolved into \textit{matrix-free} algorithms, like those embodied in matrix-free {NK--GMRES}. And through such ideas, algorithms evolved towards performing their tasks through intelligently planned \textit{recursive function evaluations}.
This leads to the premise that many scientific computations can be performed through systematic, recursive function evaluations (interspersed by brief low-dimensional tasks, like Gram-Schmidt orthogonalization, or least-squares solutions in low-dimensional subspaces) -- and thus many algorithms essentially comprise of an intelligent (recursive) concatenation of function evaluations, enhanced by some ancillary computation.
Thus, many traditional numerical algorithms \textit{are} protocols for recursively recurrent function evaluations (plus ancillary computation).
Both matrix-free {NK--GMRES} (for solving systems of nonlinear equations) and numerical initial-value problem solvers (of which RK is a notorious example) can be seen to naturally lead to \R2N2 architectures.
The analogy between an RK method and the \R2N2 was already demonstrated in our previous work \citep{Guoyue2021metalearning}.
Here, we show that the \R2N2 also represents Krylov and NK subspace methods up to the ancillary computations.

\subsubsection{Krylov subspace solvers}\label{sec:mkrylov}

Krylov subspace solvers compute an approximate solution to linear systems of the form
\begin{equation}\label{eqn:lin_solve}
    \mA\Vx=\Vb, 
\end{equation}
where $\mA \in \mathbb{R}^{m \times m}$ and $\Vb \in \mathbb{R}^m$, in an $n$-dimensional subspace $\mathcal{K}_n$, $n \leq m$, such that the norm of a residual $\Vr$, i.e., 
\begin{equation*}
    \left\Vert \Vr\right\Vert = \left\Vert\mA\Vx - \Vb \right\Vert,
\end{equation*}
is minimized \citep{Saad2003iterative}.
The Krylov subspace $\mathcal{K}_n$ is given by
\begin{equation}\label{eqn:krylov_subspace}
    \mathcal{K}_n(\mA,\Vr_0) = \spn \left\{\Vr_0, \mA\Vr_0, \mA^2 \Vr_0,\ldots,\mA^{n-1}\Vr_0\right\},
\end{equation}
and is fully determined by $\mA$ and the initial residual $\Vr_0$. In absence of prior knowledge, $\Vr_0$ is usually computed by choosing $\Vx_0 = \mathbf{0}$, i.e., $\Vr_0 = -\Vb$. 
The remaining Krylov vectors spanning the subspace are obtained by recurrent left-side multiplication of $\Vr_0$ with $\mA$. 
Finally, the approximate solution follows as
\begin{equation}\label{eqn:krylov_approx_x}
    \overline{\Vx} = \Vx_0 + \Vq^{\star}, 
\end{equation}
\begin{equation}\label{eqn:krylov_step_gmres}
    \Vq^{\star} \in \arg\min_{\Vq \in \mathcal{K}_n} \left\Vert \mA \left(\Vx_0 + \Vq\right) - \Vb \right\Vert.
\end{equation}
$\Vq^{\star}$ can be expressed as a linear combination of the vectors spanning the subspace $\mathcal{K}_n$, i.e,
\begin{equation}\label{eqn:krylov_step_lincomb}
    \Vq^{\star} = \mathbf{V}_n \bm{\upbeta} ,
\end{equation}
where $\mathbf{V}_n$ is an $n\times n$ matrix that contains the members of $\mathcal{K}_n$ as columns and $\bm{\upbeta} \in \mathbb{R}^n$ is a vector of coefficients found by \textit{solving} Problem~\eqref{eqn:krylov_step_gmres}.
Contemporary Krylov methods like \mbox{GMRES} \citep{Saad1986GMRES} orthonormalize the basis of $\mathcal{K}_n(\mA,\Vr_0)$. This operation requires additional computation but enables explicit residual minimization.

Instead of solving Equation~\eqref{eqn:krylov_step_gmres}, the \R2N2 superstructure approximates the solutions generated by the Krylov method, Equations~\eqref{eqn:krylov_subspace} -- \eqref{eqn:krylov_step_lincomb}, as follows. 
We set $\Ff(\Vx) \vcentcolon = \mA\Vx - \Vb$ such that the initial function evaluation returns $\Vv_0 = \Ff(\mathbf{0}) = \Vr_0 = -\Vb$.
Due to the nature of the layer modules $\FN_j$, Equation~\eqref{eqn:RKNN_module}, all layers after the initial layer will return outputs $\Vv_j$ that are in the span of $\left\{\Vr_0, \mA\Vr_0, \ldots, \mA^{j-1}\Vr_0 \right\}$ such that the subspace generated by the \R2N2 coincides with the Krylov subspace, Equation~\eqref{eqn:krylov_subspace}.
The output module of the \R2N2, Equation~\eqref{eqn:RKNN_final}, learns a fixed linear combination of these $\Vv_j$ through its parameters $\Vtheta_n$. By identifying these $\Vtheta_n$ as $\bm{\upbeta}$, $\Vx_k$ as $\Vx_0$ and $h=1$, one forward pass through the \R2N2 can, in principle, imitate one outer iteration of the Krylov subspace method for a single problem instance $\left(\mA,\Vb \right)$.

On the other hand, one pass through the \R2N2 is cheaper than an iteration of a Krylov method, since the \R2N2 does not perform the orthonormalization and explicit residual minimization.
Both methods require $n-1$ matrix-vector products to build the $n$-dimensional subspace, given that $\Vr_0$ is obtained for free.
Finally, we point out that some Krylov-based solvers, e.g., \mbox{GMRES}, can be iteratively restarted to compute a better approximation to a solution of a linear system \citep{Saad1986GMRES}. This restarting procedure is naturally represented by recurrent passes through the \R2N2, i.e., a restart corresponds to updating the iterate from $\Vx_k$ to $\Vx_{k+1}$.

\subsubsection{Newton-Krylov solvers}\label{sec:mNK}

Newton-Krylov solvers essentially approximate Newton iterations for the solution of a nonlinear equation system $\Ff(\Vx)=\mathbf{0}$, $\Ff \vcentcolon \mathbb{R}^m \rightarrow \mathbb{R}^m$, where the linear subproblem that arises in each Newton iteration is addressed using a Krylov subspace method \citep{Kelley1995iterative,Kelley2003solving,Knoll2004jacobian}. 
The $k$-th linear subproblem requires solving
\begin{equation*}
    \mJ(\Vx_k)\Delta \Vx_k=-\Ff(\Vx_k).
\end{equation*}
Therefore, the residual $\Vr_0$ of the linear solver corresponds to $\Ff(\Vx_k)$ and the matrix $\mA$ is substituted by the Jacobian of $\Ff$ at $\Vx_k$, $\mJ(\Vx_k) \in \mathbb{R}^{m\times m}$. The corresponding $k$-th Krylov subspace then reads
\begin{equation}\label{eqn:NK_subspace}
    \mathcal{K}_n^{(k)}\left(\mJ(\Vx_k), \Ff(\Vx_k)\right) = \spn \left\{\Vv_0, \ldots, \Vv_{n-1}, \right\},
\end{equation}
\begin{equation*}
    \Vv_j = \mJ(\Vx_k)^j \Ff(\Vx_k) \quad \forall \, j \in \{0,\ldots,n-1\}.
\end{equation*}
In the practical use-case of Newton-Krylov methods, $\mJ(\Vx_k)$ is not computed explicitly. Instead, matrix-vector products $\mJ(\Vx_k) \Vz$, where $\Vz \in \mathbb{R}^m$ is a vector, are approximated using a directional derivative of $\Ff$ at $\Vx_k$ \citep{Kelley2003solving}, i.e., 
\begin{equation*}
    \mJ(\Vx_k) \Vz \approx \frac{\Ff(\Vx_k +\epsilon \Vz) - \Ff(\Vx_k)}{\epsilon},
\end{equation*}
where $\epsilon$ is a small value in the order of $10^{-8}$. 
The RHS corresponds to forward-differencing, which is shown to lie within the \R2N2 superstructure in the supplementary material (\textbf{SM1}). 
Typically, multiple iterates $\Vx_k$ have to be computed to sufficiently approximate a solution of $\Ff(\Vx)=\mathbf{0}$.
Such iterative behavior can be reproduced by performing multiple recurrent passes through the \R2N2.
Since $\mJ(\Vx_k)$ is always computed with the current iterate $\Vx_k$, each pass through the \R2N2 necessarily corresponds to a new Newton iteration if the remaining identities noted in the previous section, \Cref{sec:mkrylov}, on linear Krylov solvers are applied again. 
Consequently, the limited expressivity due to $\Vtheta_n$ as opposed to the minimization in a Krylov method is inherited, too.

\subsection{Training the \R2N2 superstructure}\label{sec:mloss}

Training the \R2N2 superstructure implies solving Problem~\eqref{eqn:generic_learningproblem} for the trainable parameters $\Vtheta$.
The input data for each problem class is sampled from a distribution representing a set of problem instances, the solutions to which are the training targets. 
In the following, we indicate data samples by an additional subscript $i=1,\ldots,N$, where $N$ is the total number of samples, i.e., $\Vx_{i,k}$ refers to the $i$-th sample after the $k$-th iteration.
The output of the $k$-th pass through an \R2N2 is the iterate denoted by $\hat{\Vx}_{i,k+1}$.
As a loss function we use a weighted version of the mean squared error (\textit{MSE}) between $\hat{\Vx}_{i,k}$ and $\Vx_{i,k}^{target}$ or the corresponding residual $\Pi\left(\hat{\Vx}_{i,k},F_i\right)$, summed over all iterations, i.e.,
\begin{subequations}\label{eqn:MSE_losses} 
\begin{equation}\label{eqn:MSEloss_x}
    MSE_x=\frac{1}{N} \sum_{k=1}^T \sum_{i=1}^N w_{i,k} {\left\|\hat{\Vx}_{i,k} - \Vx^{target}_{i,k}\right\|_2}^2,
\end{equation}
\begin{equation}\label{eqn:MSEloss_fx}
    MSE_{\Pi}=\frac{1}{N} \sum_{k=1}^T \sum_{i=1}^N w_{i,k} {\Pi(\hat{\Vx}_{i,k},F_i)}^2,
\end{equation}
\end{subequations}
where $w_{i,k}$ are the weights belonging to the $i$-th sample in the $k$-th iteration.
Whether we utilize Equation~\eqref{eqn:MSEloss_x} or Equation~\eqref{eqn:MSEloss_fx} is problem-specific. For instance, for equation solvers, we can make use of $\Ff(\Vx^\star_i):=\mathbf{0}$ to compute the residual for Equation~\eqref{eqn:MSEloss_fx} from $\Ff(\Vx_{i,k+1})$. 
For integrators on the other hand, we can sample training targets $\Vx_{i,k+1}^{target}$ from the trajectory computed by some high-order integrator or, if available, use an analytic solution to generate target data.
We do not use any regularizers for training the \R2N2 in this work.

\subsection{Implementation}\label{sec:mimplementation}

We implemented the \R2N2 superstructure in PyTorch (version $1.8.0$) \citep{Paszke2019pytorch}.
We trained on an Intel i7-9700K CPU using \textit{Adam} \citep{Kingma2017adam} for $25,000$ epochs with the default learning rate of $0.001$ or L-BFGS \citep{Liu1989lbfgs} for $5,000$ epochs with a learning rate of $0.01$. 
Note that L-BFGS is commonly used to fine-tune networks with comparable architecture as ours that were pretrained with \textit{Adam}, e.g., by \cite{Zhao2020discovery}, suggesting that L-BFGS can improve the training result.
A detailed assessment of the capability of different optimizers and strategies to train the \R2N2 superstructure is beyond the scope of this work.

%% file: sections_rev/04_Results_Lin.tex
\section{Numerical experiments}\label{sec:results}

In this section, we demonstrate the ability of the \R2N2 superstructure to learn efficient iterations for both equation solvers and integrators. 
\Cref{sec:rnonlinear} and \Cref{sec:rint} will show that computational benefits of solvers trained for a specific problem class start to arise in solving \textit{nonlinear} problems with such algorithms.
In particular, \Cref{sec:rnonlinear} demonstrates the extended applicability of the \mbox{RK-NN} introduced in \cite{Guoyue2021metalearning} to \textit{nonlinear} equation systems.
We start here, however, with \textit{linear} systems of equations in \Cref{sec:rlinear}, not because of the computational benefits achievable, but because the comparison between the \R2N2 and GMRES (as a matrix-free linear algebra solver) facilitates initial insight into the functionality of the \R2N2 when learning iterative solvers for equation systems.

In all experiments that follow, the \R2N2 and the classical method it is compared to are allowed the same number of function evaluations per iteration. Thus, for equation solvers, the \R2N2 requires less overall operations per iteration, c.f. \Cref{sec:mcoefficient_matching},
and for integrators, the amount of overall operations per iteration will be identical.
We have consistently used $70\%$ of the generated data for training and an independent sample of $30\%$ for the test results that are presented in the following. 

\subsection{Solving linear equation systems}\label{sec:rlinear}

First, we study the solution of linear equations, see Equation~\eqref{eqn:lin_solve}.
For the illustrative experiments we consider examples where a single task is given by $F_i=\left(\mA_1,\Vb_i\right)$ with $\mA_1\in \mathbb{R}^{m\times m}$, $\Vb_i \in\mathbb{R}^{m}$, $\Vx\in \mathbb{R}^{m}$ and $m=5$.
$\mA_1$ is a fixed, randomly-generated symmetric positive definite matrix overlaid with an additional boost to its diagonal entries to influence its spectrum. Right-hand sides (RHS) $\Vb_i$ are sampled uniformly around a fixed randomly-chosen mean. See supplementary material (\textbf{SM2}) for details.
As training input we use the data set of tasks $\{F_i\} = \{(\mA_1, \Vb_i)\}$ and always select $\Vx_0 = \mathbf{0}$ as the initial point. Therefore, the resulting $n$-dimensional Krylov subspace is always spanned by $\left\{\Vb_i,\mA_1 \Vb_i,\ldots, \mA_1^{n-1}\Vb_i\right\}$. 
When injecting the input data to the \R2N2, we resort to $\Ff(\Vx_i)\vcentcolon= \mA_1\Vx_i-\Vb_i$.
We use $\Ff\left(\Vx_{i,k+1}^t\right) = \mathbf{0}$ as training targets for all samples $i$ and all steps $k$. 
Therefore, the training loss at a step $k$, derived from Equation~\eqref{eqn:MSEloss_fx}, becomes
\begin{equation}\label{eqn:linsolve_loss}
    MSE_{f,k}= \sum_{k=1}^T w_k \sum_{i=1}^N \left(\mA_1\hat{\Vx}_{i,k}-\Vb_i \right)^2,
\end{equation}
where $w_k$ becomes relevant when training over multiple iterations, $T > 1$, and is tuned by hand. 

To evaluate the performance of the \R2N2, we compute the reduction of the norm of the residual that was achieved after $k$ iterations through the \R2N2, i.e.,
\begin{equation}\label{eqn:res_reduction}
    \Delta \Vr_{i,k} = \left\Vert \Vb_i \right\Vert - \left\Vert \mA_1\hat{\Vx}_{i,k}-\Vb_i \right\Vert.
\end{equation}
For the results that follow, we compare the \R2N2 to the SciPy implementation of the solver GMRES \citep{Saad1986GMRES, Scipy2020}. 
GMRES is set to use the same number of inner iterations, i.e., function evaluations as the \R2N2.
Considering the restarted version of GMRES, GMRES(r), a forward pass through the \R2N2 represents one outer iteration \citep{Saad1986GMRES}. 
We evaluate the reduction in residual norm divided by the one achieved using GMRES, i.e., $\frac{\left\Vert \Delta \Vr_{1,i} \right\Vert_{R2N2}}{\left\Vert \Delta \Vr_{1,i} \right\Vert_{GMRES}}$, with the subscripts `R2N2' indicating the \R2N2 and `GMRES' indicating the solver GMRES. 

\begin{figure*}[t!] 
    \centering
    \subcaptionbox{Training and testing with only $\mA_1$. Light dots show training samples. \label{fig:linear_results_singleA}}[.48\textwidth]{\includegraphics[scale=0.98]{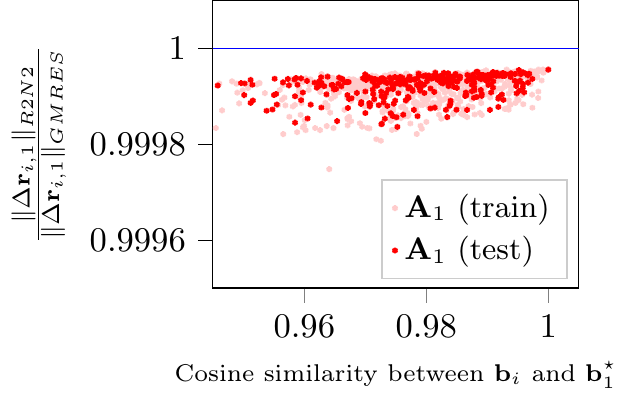}}
    \hfill
    \subcaptionbox{Testing on instances featuring $\mA_1$, $\mA_2$ and $\mA_3$. \label{fig:linear_results_multiA}}[.48\textwidth]{\includegraphics[scale=0.98]{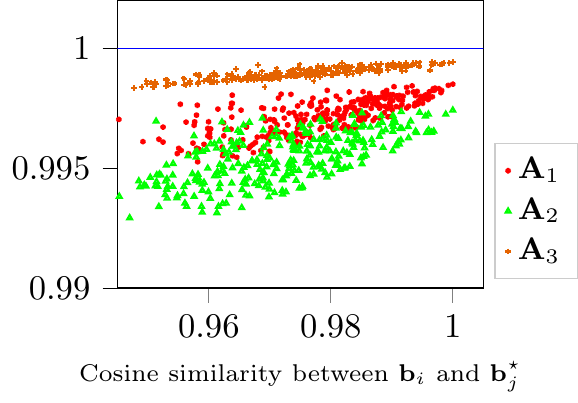}}
    \caption{Relative performance of \R2N2 vs GMRES with $n=4$ inner iterations for problem instances of Problem~\eqref{eqn:lin_solve}. Each dot indicates a different RHS $\Vb_i$. The blue lines indicate the baseline achieved by GMRES. Subscripts `R2N2' and `GMRES' stand for the \R2N2 and GMRES, respectively. Each method returns a step in a $4$-dimensional subspace, which is generated via $3$ function evaluations.
    The residual reduction is computed using Equation~\eqref{eqn:res_reduction} with the respective $\mA_j$ for $j \in \{1,2,3\}$. The scales in a) and b) are different.
    By $\Vb_j^{\star}$ we denote the sample for which the relative performance of \R2N2 is maximized for each case. 
    In (a), the training data set features only problems with $\mA_1$, while in (b), three different matrices are used.}
    \label{fig:linear_results}
\end{figure*}

The training result of this first experiment is shown in \Cref{fig:linear_results_singleA}. 
Notably, the performance of the \R2N2 is upper-bounded by the performance of GMRES, as {GMRES} minimizes the residual in the subspace spanned by the Krylov vectors.
Given that this subspace is invariant with respect to the operations that can be learned by the layers of the \R2N2 for the linear problem instances, the \R2N2 cannot improve on the performance of GMRES in this first case study.
The experiment with a fixed $\mA_1$ was chosen to illustrate this upper bound of the \R2N2 and does not yield a proper problem distribution for the training set -- with $\mA_1$ fixed, only a mapping from $\Vb_i$ to $\Vx_i^{\star}$ needs to be learned.
Thus, we now draw two additional matrices, $\mA_2, \mA_3 \in \mathbb{R}^{m\times m}$ from the distribution (see (\textbf{SM2})). Through the addition of $\mA_2$ and $\mA_3$, the performance, i.e., the adaptation, of the resulting \R2N2 on problem instances formed with either of the three matrices is decreased compared to the previous case (see \Cref{fig:linear_results_multiA}). 

\begin{figure*}[htb] 
    \pgfplotsset{/pgfplots/group/.cd,
        horizontal sep=0.5cm,
        vertical sep=0.5cm
    }
    \centering
    \includegraphics[scale=1]{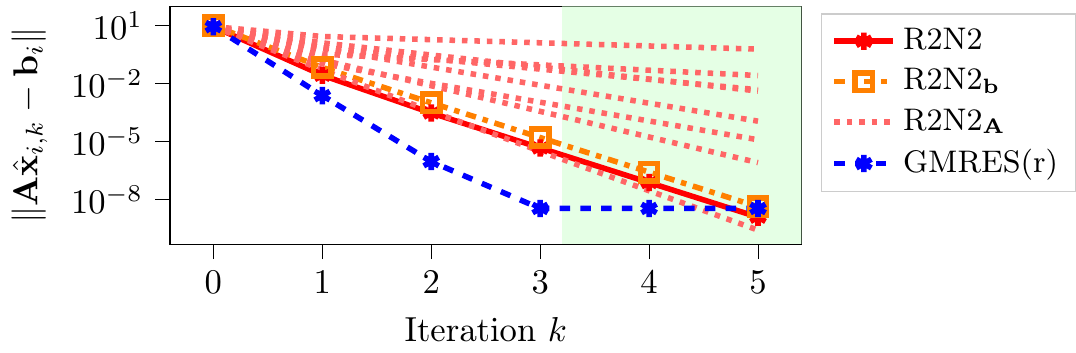}
    \caption{Convergence of the \R2N2 vs GMRES(r) for solving \eqref{eqn:lin_solve} for multiple samples of $\Vb_i$. The lines connect the sample mean at each (outer) iteration $k$. Both solvers use $n=4$ inner iterations. The \R2N2 was trained on $T=3$ outer iterations, using Equation~\eqref{eqn:linsolve_loss} with $w_k=4^k$ as a loss, and then applied over $5$ iterations. Iterations within the shaded square area, $k>3$, have not been seen in training and indicate one type of extrapolation. 
    The orange-squared line `\R2N2$_{\Vb}$' shows extrapolation in the RHS, and the dotted lines `\R2N2$_{\mA}$' show extrapolations in matrices $\mA$ (see (\textbf{SM3.1}) for details).}
    \label{fig:lin_singleparam_conv}
\end{figure*}

Next, we study the performance of the \R2N2 over multiple iterations together with the extrapolation capabilities of the \R2N2. 
Here, the \R2N2 is compared with the restarted version of GMRES, \textit{GMRES(r)}, where each iteration uses the output of the previous iteration as an initial value. 
We train the \R2N2 to minimize loss after three outer iterations, given by Equation~\eqref{eqn:linsolve_loss}. We set $w_k=4^k$, which was found to yield decent results.
The training dataset is again generated from the random draw of samples of $\Vb_i$ combined with the three matrices $\mA_1$, $\mA_2$, and $\mA_3$.
The solid red line in \Cref{fig:lin_singleparam_conv} shows the convergence of the \R2N2 by plotting the average residual of the test set for $\mA_1$ only (for clarity). 
The \R2N2 reduces the residual over all consecutive outer iterations, i.e., even beyond the third outer iteration, which indicates a first type of successful extrapolation of the \R2N2: Iterates $\Vx_k$ for $k > 3$ have not been contained by the training input distribution, yet the \R2N2 iterations progress towards the solution of the problem beyond that point.
Related to this extrapolation on RHS given by iterates $\Vx_k$ is the orange dash-dotted line that examines the \R2N2 on uniform random RHSs $\Vb_i$, normalized to the length of the test set samples. 
The \R2N2 is shown to converge to a solution for \textit{all} of these RHSs, i.e., the convergence is due to a property of the \R2N2 and $\mA_1$. We analyze this further in supplementary material (\textbf{SM4}).
Finally, we analyze two additional types of extrapolation applied to $\mA_1$: i) raising the noise level of its random component by up to a factor of $7$, ii) reducing or increasing the induced spectrum in $\mA_1$, respectively. See supplementary material (\textbf{SM3.1}) for details.
Trajectories for the resulting test matrices -- again combined with the initial RHS samples of $\Vb_i$ -- are plotted as light-red dotted lines in \Cref{fig:lin_singleparam_conv}. The results demonstrate that the \R2N2 learns a solver adapted to a problem data set, and, further, that the \R2N2 extrapolates reasonably well beyond that data set, with a performance decrease as $\mA$ becomes more different from the training distribution. 
The extrapolation experiments are presented in more detail -- including some failure cases -- in (\textbf{SM3.1}). 
We also show that vanilla neural networks (NNs) of comparable size cannot learn a solver like the \R2N2 (\textbf{SM3.2}),
and that a \R2N2 modified to \textit{predict} solutions does better than the NNs in this task too (\textbf{SM3.3}).

\begin{figure*}[htb]
    \pgfplotsset{/pgfplots/group/.cd,
        horizontal sep=0.5cm,
        vertical sep=0.5cm
    }
    \centering
    \includegraphics[scale=1]{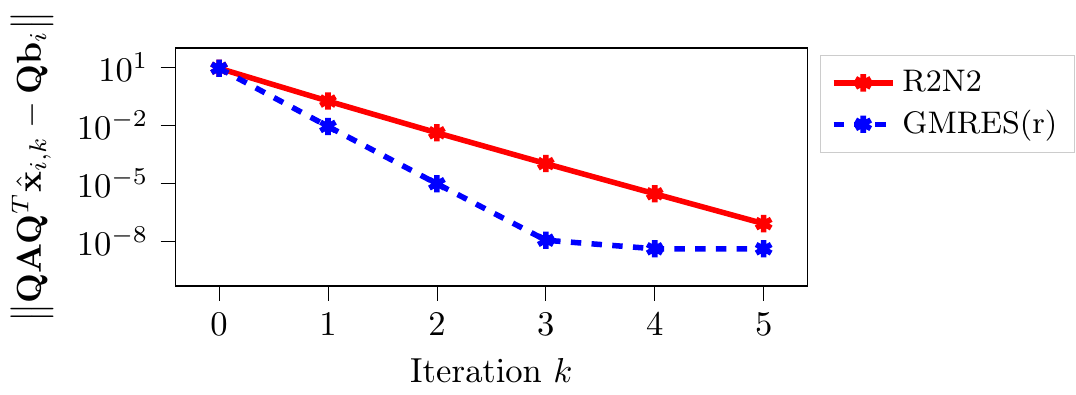}
    \caption{Convergence of the \R2N2 applied to the embedded problem \eqref{prob:lin_lifted}. The \R2N2 was trained on samples from the original problem \eqref{eqn:lin_solve} (same \R2N2 as the one used in \Cref{fig:lin_singleparam_conv}). Both solvers again use $n=4$ inner iterations.
    The lines connect the sample mean at each (outer) iteration $k$.}
    \label{fig:lin_conv_LIFTED}
\end{figure*}

As a final example, we demonstrate that the \R2N2 can also be applied to an embedding of the problem class.
We generate a random $15$-dimensional orthonormal matrix $\mathbf{Q}$ from the Haar distribution via SciPy \citep{Scipy2020, Mezzadri2006generate}, and consider the problem class with instances 
\begin{equation}\label{prob:lin_lifted}
    \mathbf{Q} \mA_1 \mathbf{Q}^T \Vx = \mathbf{Q}\Vb_i,
\end{equation}
where by abuse of notation $\mA_1$ and $\Vb_i$ are made $15$-dimensional by zero-padding the new dimensions.
The convergence of the residual of this embedded problem is shown in \Cref{fig:lin_conv_LIFTED}.
Evidently, as an adapted solver, the \R2N2 is also applicable to this embedding of the problem class it was trained on.
By inspecting the embedding, we see that the resulting subspace vectors are also rotated by $\mathbf{Q}$, i.e.,
\begin{equation*}
    \mathcal{K}_n\left(\mathbf{Q}\mA\mathbf{Q}^T, \mathbf{Q}\Vb\right) = \left(\mathbf{Q}\Vb, \mathbf{Q}\mA\Vb, \ldots, \mathbf{Q}\mA^{n-1}\Vb \right),
\end{equation*}
as the $\mathbf{Q}^T\mathbf{Q}$ in the inner of the matrix-vector products cancels out each time.

%% file: sections_rev/04_Results_NK.tex
\subsection{Solving nonlinear equation systems}\label{sec:rnonlinear}

As an illustrative nonlinear equation system, we consider Chandrasekhar's $H$-function in conservative form, an example that was extensively used in Kelley's book on Newton-Krylov solvers \citep{Kelley2003solving}.
The discretized form of this equation reads
\begin{equation}\label{eqn:chand_H_disc}
    \left(\Ff(\Vx)\right)_j = \Vx_j - \left(1 - (\mA_c\Vx)_j \right)^{-1},
\end{equation}
for $j=1, \ldots, m$ with the entries of parametric matrix $\mA_c \in \mathbb{R}^{m \times m}$ defined by
\begin{equation*}
   A_{c,j\iota} = \frac{c \mu_j}{2m \left(\mu_j+\mu_\iota\right)},
\end{equation*}
with $\mu_\iota = \frac{(\iota-\frac{1}{2})}{m}$ and $\iota=1, \ldots, m$ also indexing the discretization points.
The derivation of the equations is given in supplementary material (\textbf{SM2}).
We combine problems with $m=10$ and $m=20$ discretization points and $c \in \left\{0.875,0.905,0.935\right\}$ to span a training set.
For all of these problems, we sample initial values $\Vx_{0,i}$ from a normal distribution with mean $\mathbbm{1}^{(m)}$, where $\mathbbm{1}^{(m)}$ is an $m$-dimensional vector of ones, and variance $0.2\cdot\mathbbm{1}^{(m)}$.
We again use \eqref{eqn:linsolve_loss} as a loss function, where targets are $\Ff\left(\Vx_{k+1,i}^t\right)=0$ for all samples $i$ and iterations $k$.
The SciPy implementation of Kelley's Newton-Krylov GMRES (NK-GMRES) serves as a benchmark \citep{Scipy2020, Kelley2003solving}.

\begin{figure*}[t!] 
    \centering
    \subcaptionbox{After $k=1$ nonlinear iterations. \label{fig:nonlinear_multif_31}}[.48\textwidth]{\includegraphics[scale=0.97]{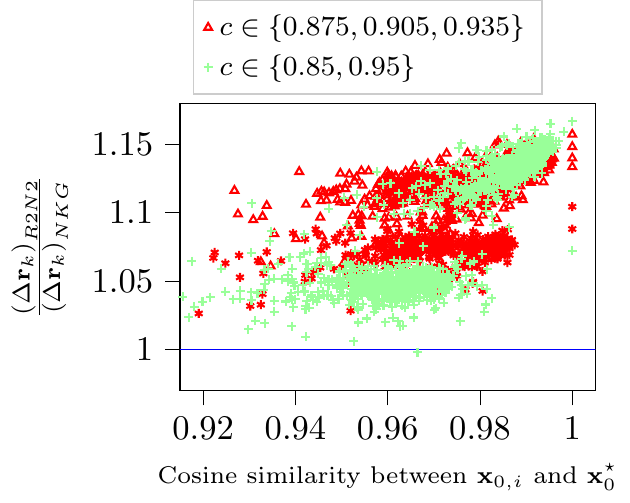}}
    \hfill 
    \subcaptionbox{After $k=2$ nonlinear iterations. \label{fig:nonlinear_multif_32}}[.48\textwidth]{\includegraphics[scale=0.97]{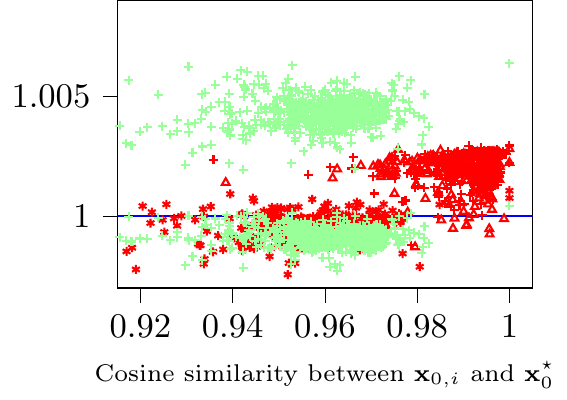}}
    \caption{Relative performance of \R2N2 vs NK-GMRES using $n=3$ inner iterations and one 
    (a) or two (b) outer iterations on Problem~\eqref{eqn:chand_H_disc} with various coefficients $c$ and various initial guesses $\Vx_{0,i}$. 
    Note the differences in scales between the two subfigures. 
    Instances where $c$ is extrapolated outside of the training range are plotted in light green.
    The residuals are computed using Equation~\eqref{eqn:res_NK} with subscripts `R2N2' and 'NKG' for the \R2N2 and the NK-GMRES solver, respectively. $\Vx_0^{\star}$ refers to the sample for which the relative performance of \R2N2 is best in the respective experiment.}
    \label{fig:nonlinear_multif}
\end{figure*}

\begin{figure*}[htb]
    \pgfplotsset{/pgfplots/group/.cd,
        horizontal sep=0.5cm,
        vertical sep=0.5cm
    }
    \centering
    \includegraphics[scale=1]{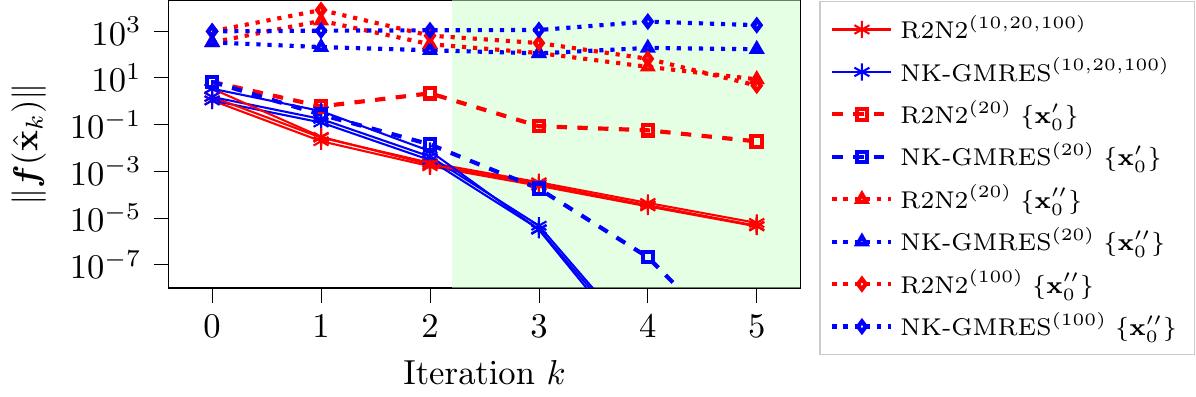}
    \caption{Convergence of the \R2N2 (red) vs NK-GMRES (blue) for $8$ nonlinear iterations of solving \eqref{eqn:chand_H_disc}. Both solvers use $3$ inner iterations. The \R2N2 used herein coincides with the one shown in \Cref{fig:nonlinear_multif_32}, i.e., it was trained only on the first $2$ nonlinear iterations. 
    The shaded area $k>2$ denotes performing more iterations than were used for training.
    The superscripts indicate the set of problem dimensions $m$ used.
    Performance for samples of $\{\Vx_0\}$ from the training range are shown by solid lines. Dashed lines use initial samples $\{\Vx'_0\}$ with 10-fold noise around $\mathbbm{1}^{(m)}$, whereas dotted lines use initial samples $\{\Vx''_0\}$ centered at $5 \cdot\mathbbm{1}^{(m)}$.}
    \label{fig:NK_conv}
\end{figure*}

We evaluate the performance after $k$ nonlinear iterations by the reduction in the norm of the residual:
\begin{equation}\label{eqn:res_NK}
    \Delta \Vr_k = \left\Vert \Ff(\Vx_{0}) \right\Vert - \left\Vert \Ff(\Vx_{k}) \right\Vert.
\end{equation}
\Cref{fig:nonlinear_multif} showcases the performance of the \R2N2 on individual samples after training for $2$ nonlinear iterations. The \R2N2 is able to achieve more progress within these first two iterations than NK-GMRES even though it does not explicitly perform the minimization in the Krylov subspace to compute the step. The advantage in performance is maintained over a range of values for coefficient $c$ in $\mA_c$ both within the training range (red markers) and outside of the training range (light green). 
We therefore deduce that the \R2N2 has learned to construct a subspace that contains more of the true solution for a specific problem class than the subspace formed by NK-GMRES. 
This is possible because the subspace $\mathcal{K}_n\left(\mJ(\Vx_k),-\Ff(\Vx_k)\right)$, Equation~\eqref{eqn:NK_subspace}, used in the $k$-th iteration depends not only on $\left(\Ff,\Vx_k\right)$ but also on the trainable parameters $\Vtheta_j$ in the $\FN_j$ modules of the \R2N2, cf. Equation~\eqref{eqn:RKK_member_j}. Finally, we observed no notable difference between problem samples with $m=10$ and $m=20$.
We applied the \R2N2 from \Cref{fig:nonlinear_multif} for a total of $7$ nonlinear iterations, see \Cref{fig:NK_conv}. The advantage of the \R2N2 over NK-GMRES vanishes for $3$ and more nonlinear iterations, but the \R2N2 does still converge to an approximate solution when applied iteratively.
Note, however, that this promising finding does not guarantee that the \R2N2 can be trained to converge to a solution of arbitrary nonlinear equation systems.
Furthermore, \Cref{fig:NK_conv} shows extrapolation to problems with $m=100$. Similar to the case in the previous subsection, the \R2N2 is agnostic of the problem dimension, which in this case allows generalization across various discretizations of the problem. 
Thus, overall the \R2N2 is applicable to solving instances of Equation~\eqref{eqn:chand_H_disc} without explicit access to either of its generative properties, i.e., $c$ and $m$, something an unstructured neural network trained on problem data was found to be incapable of, cf. (\textbf{SM2}). 
Finally, we studied extrapolation for different initial guesses. The test set denoted by $\{\Vx'_0\}$ is generated with 10-fold variance compared to the training set, whereas the test set denoted $\{\Vx''_0\}$ consists of initial guesses centered around a different vector, here $5\cdot\mathbbm{1}^m$s. For the latter, the \R2N2 shows some convergence, also continuing over further iterations that were not plotted herein. After 15 iterations, all test samples except for a couple (that are still converging) have reached a tolerance of $1e-8$. 
NK-GMRES, on the other hand, produces more outliers: after 15 iterations, only $63.2~\%$ of samples have converged. This number is increased to $76.7~\%$ after 20 iterations, $98~\%$ after 50 iterations and $99.4~\%$ after 75 iterations. That is, the mean curves are dominated by the instances that are slow to converge. 
This distinct difference in performance between the \R2N2 and NK-GMES merits further study.

%% file: sections_rev/04_Results_RK.tex
\subsection{Solving initial-value problems}\label{sec:rint}

Finally, we study the solution of initial-value problems with the known RHS $\Ff$, i.e., Problem~\eqref{eqn:integration_fp}. 
We covered learning integrators thoroughly in our previous work \citep{Guoyue2021metalearning} where we employed a Taylor series-based regularization to promote certain orders of convergence as property of the \mbox{RK-NN}.
We now demonstrate that \mbox{RK-NN} integrators that outperform classical RK integrators can also be learned without special regularizers and, as a further extension of our previous work, that the \mbox{RK-NN} integrators work over multiple timesteps.
All \mbox{RK-NNs} in this section are trained from the \R2N2 superstructure and we thus name them \R2N2 in the remainder of the section.

We reiterate the van der Pol oscillator from our previous work, i.e., 
\begin{subequations}\label{eqn:vdP}
\begin{align}
    \dot{x}_{(1)}(t) & =  x_{(2)}(t), \\
    \dot{x}_{(2)}(t) & =  a\left(1-x_{(1)}^2(t)\right)x_{(2)}(t) - x_{(1)}(t),
\end{align}
\end{subequations}
with $\Vx=\left(x_{(1)},x_{(2)}\right)$.
For data generation, we sample coefficients $a \sim \mathcal{U}(1.35,1.65)$, initial values $x_{(1)}(t=t_0)  \sim \mathcal{U}(-4,-3)$ and $x_{(2)}(t=t_0)  \sim \mathcal{U}(0,2)$ and timesteps $h \in \left[0.01,0.1\right]$ equidistantly
Note that $\Vx(t=t_0)$ and $h$ are contained in the problem parameters $\Vp$. For Problem~\eqref{eqn:integration_fp}, we use them directly as the inputs $\Vx_k$ and $h$ of the \R2N2, cf. \Cref{fig:superstruct}.
We generate target data (that also servers as ground truth) for the timesteps $h$ using SciPy's \textit{odeint} \citep{Scipy2020} with error tolerance set to $10^{-8}$.
Training loss is calculated by the following specification of Equation~\eqref{eqn:MSEloss_x}: 
\begin{equation*}
    MSE_x=\sum_{k=1}^T \sum_{i=1}^N \frac{\left( \hat{\Vx}_{i,k} - \Vx_{i,k}^t \right)^2}{h_i^p}.
\end{equation*}
The losses are summed over $T$ integration steps of a trajectory with the timestep $h$, i.e., at times $t_0+h, t_0+2h, \ldots, t_0+Th$.
Further, the denominator of the loss terms allows weighting samples based on the timestep of a specific sample, $h_i$, and, in particular, weighting according to an expected convergence order $p$. 
We set $p=n$ for the results in \Cref{sec:rint}, where $n$ is the number of layers of the RK-NN.
The function evaluations in the \R2N2 directly evaluate the RHS of Equation~\eqref{eqn:vdP}. One pass through the \R2N2 therefore resembles one step of a RK method.

\begin{figure*}[t!] 
    \centering
    \subcaptionbox {The \R2N2 error after one step is shown. \label{fig:integration_results_multif_1step}}[.48\textwidth]{\includegraphics[scale=1]{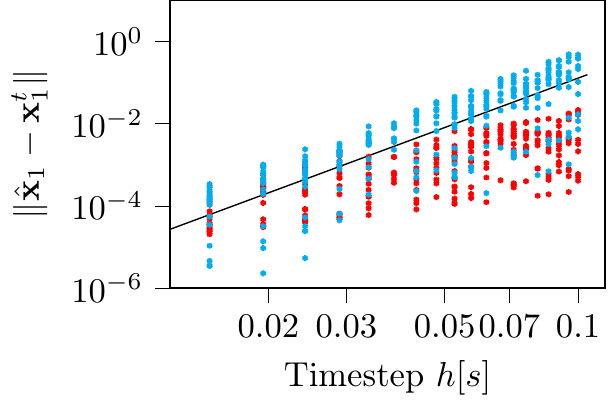}}
    \hfill
    \subcaptionbox{The \R2N2 error after two steps is shown. \label{fig:integration_results_multif_2step}}[.48\textwidth]{\includegraphics[scale=1]{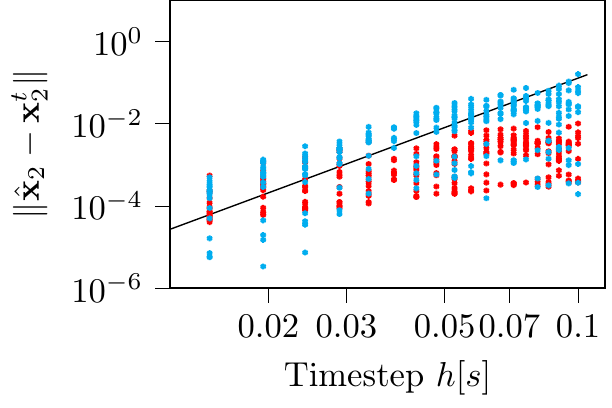}}
    \par
    \subcaptionbox{The \R2N2 error after three steps is shown. \label{fig:integration_results_multif_3step}}[.48\textwidth]{\includegraphics[scale=1]{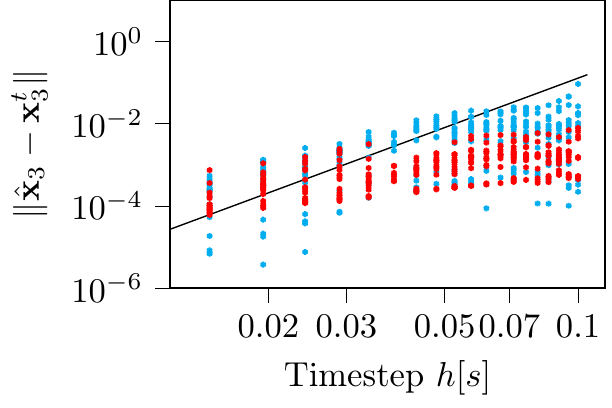}}
    \hfill
    \subcaptionbox{The \R2N2 error after five steps is shown. \label{fig:integration_results_multif_5step}}[.48\textwidth]{\includegraphics[scale=1]{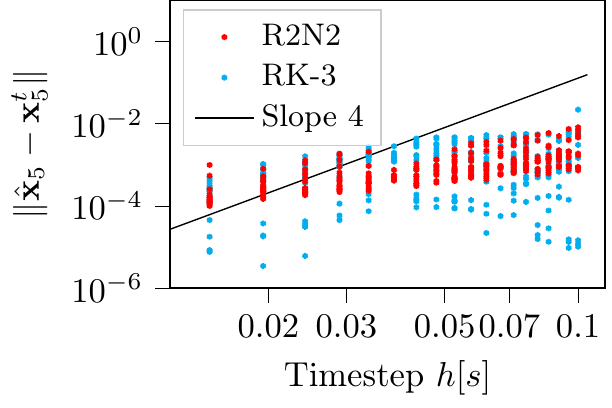}}
    \caption{Performance of \R2N2 vs \mbox{RK-$3$} integrating the van der Pol oscillator equations, Equation~\eqref{eqn:vdP}, with varying coefficients $a$, stepsizes $h_i$ and initial conditions $\Vx_{0,i}$. The figures show the same \R2N2, trained on one timestep, and evaluated after one (a), two (b), three (c) and five timesteps (d), respectively. The slope of 4 is added to indicate the nominal local truncation error of RK-3.}
    \label{fig:integration_results_multif}
\end{figure*}

To assess the performance of the learned integrators, we compare the \R2N2 instantiated with $n$ layers to classical RK methods of $n$ stages, denoted by \mbox{RK-$n$}.
The training data contains samples with varying coefficients $a_i$, timesteps $h_i$ and initial values $\Vx_i(t=0)$.
The error for either integrator is evaluated against a ground truth approximated by \textit{odeint} and denoted $\Vx_k^t$.
Results using $n=3$ on the van der Pol oscillator, Equation~\eqref{eqn:vdP}, are shown in \Cref{fig:integration_results_multif}. 
In \Cref{fig:integration_results_multif_1step}--\ref{fig:integration_results_multif_5step}, we show results for training on $T=1$ step of integration and evaluating on $T=1$, $T=2$, $T=3$, and $T=5$ steps, respectively, i.e., for $k>1$, the iterates $\Vx_k$ that go into the \R2N2 as inputs lie outside of its training range. We thus test for, and verify, some limited generalization.
Over the first three timesteps, the \R2N2 integrates the van der Pol oscillator equations more accurately than \mbox{RK-$3$}.
We want to stress here that the classical RK method uses fixed coefficients for the computation of stage values and for the computation of the step itself.
Therefore, the \R2N2 can improve over the classical RK with regards to both of these computational steps. 
Evidently, we are able to learn coefficients $\Vtheta_j$ in the stages $\FN_j$ and $\Vtheta_n$ for the output $\FN_n$ that lead to a better approximation of the integral for problems from the training distribution, including limited generalization, than the RK method does.
 
Overall, we find that the accuracy of the \R2N2 gradually decreases over the number of timesteps such that the \R2N2 is not more accurate than RK-$3$ after the $5$-th iteration anymore.
The trend observed in \Cref{fig:integration_results_multif_1step}--\ref{fig:integration_results_multif_5step}, although less pronounced, is consistent with the behavior of the \R2N2 over multiple iterations of equation solving.

%% file: sections_rev/05_Conclusion.tex
\section{Conclusion}\label{sec:conc}

This work proposes an alternative, augmented perspective for the use of the \mbox{RK-NN}, a recurrent neural network templated on RK integrators, as a recursively recurrent superstructure for a wider class of iterative numerical algorithms.
The \R2N2 superstructure embeds function evaluations inside its layers and feeds the output to all successive layers via forward skip connections and linear combinations. The embedding of function calls into the architecture disentangles the algorithm to be learned from the function it acts on.
We have shown that the \R2N2 superstructure provides an inductive bias towards iterative algorithms based on recurrent function evaluation, e.g., the well-established Krylov subspace solvers and RK methods. 
Our numerical experiments demonstrate that, thus, the \R2N2 superstructure can mimic steps of Krylov, Newton-Krylov and RK algorithms, respectively.

In particular, when learning a single step of linear equation solvers (\Cref{sec:rlinear}), the performance of the \R2N2 is bounded by that of the benchmark GMRES. This is consequential considering that the subspaces generated by the two approaches are equal but GMRES further \textit{minimizes} the residual in this subspace, whereas the \R2N2 learns a linear combination that can coincide with the minimizer, cf. Equation~\eqref{eqn:krylov_step_gmres}, for at most a single problem instance. 
In contrast, a nonlinear equation solver (\Cref{sec:rnonlinear}) can improve upon iterations performed by NK-GMRES, because in this case the \R2N2 can learn to construct subspaces in which the residual is reduced more strongly than by NK-GMRES. 
Applying the \R2N2 for multiple outer iterations resembles a restarted iterative solver for linear problems or a Newton-Krylov method for nonlinear problems, respectively.
We have empirically demonstrated the ability of the \R2N2 to converge to solutions of linear and nonlinear equation solving problems, however, we cannot guarantee the convergence of trained solvers for arbitrary problems.
The \R2N2 is also capable of extrapolation to similar problems as the ones seen during training. This includes not only extensions to the range of problem parameters but also embedding in higher dimensions or finer discretization.
Finally, we have revisited our previous work about learning RK integrators \citep{Guoyue2021metalearning} by demonstrating successful integration over multiple timesteps (\Cref{sec:rint}). 
Our results suggest that the advantage of the \R2N2 or \mbox{RK-NN}, respectively, over a classical RK method cannot be sustained over longer time horizons of integration.

In summary, iterative algorithms trained within the \R2N2 superstructure can, when possible, find a subspace in which the residual can be reduced more than by their classical counterparts, given the same number of function evaluations.

%% file: sections_rev/06_Outlook.tex
\section{Future research directions}\label{sec:outlook}

\subsection{Application to other computational problems}\label{sec:oappl}

For further application, the compatibility of the \R2N2 superstructure with other problem classes that comply with the general form of Problem~\eqref{eqn:generic_task} and that are solved with iterative algorithms, e.g., eigenvalue computation, PCA decomposition \citep{Gemp2020eigengame}, or computation of Neumann series \citep{Liao2018reviving}, could be assessed. 
Moreover, besides learning an algorithm as a neural network architecture that maps from a set of problems to their solutions, the superstructure proposed in this work can also be deployed in the \textit{inverse} setting, i.e., to identify a problem or function under the action of the known algorithm given input/output data.
This was the original motivation in \cite{RicoMartinez1995nonlinear}, leading to nonlinear system identification, that can be analyzed using inverse backward error analysis \citep{Zhu2020inverse,Zhu2023implementation}.

\subsection{Extension of the superstructure}\label{sec:oext}

Other, more intricate algorithms can be represented by the proposed superstructure if its architecture is extended with additional trainable modules.
For instance, \textit{general linear methods} for integration (see \cite{Butcher.2016}) can be captured by the superstructure if not just the final output is subject to recurrence, but the layer outputs $\Vv_j$ are too (cf. with \Cref{fig:superstruct}).
Moreover, a preconditioner (see Section~8 of \cite{Saad2000iterative}) can be inserted inside the layers of the superstructure: for instance, a cheap, approximate model of $\Ff$ can be used to to effectively build such a preconditioner \citep{Qiao2006spatially}.
Finally, non-differentiable operations that are part of most algorithms, e.g., the checking of an error tolerance, can be included via smoothed relaxations \citep{Ying2018hierarchical,Tang2020towards}.

\subsection{Neural architecture search}\label{sec:oNAS}

Larger architectures will eventually call for superstructure optimization to yield parsimonious algorithms, i.e., determining the optimal \mbox{(sub-)}structure for a given set of problem instances together with the optimized parameters. This challenge can be formulated as a mixed-integer nonlinear program (MINLP).
Presently, these problems are addressed by heuristic methods, referred to as neural architecture search (NAS, \cite{Elsken2019neural, Hospedales2020metalearning}), that are relatively efficient in finding good architectures.
For the superstructure, three types of NAS methods appear suited: i)~structured search spaces to exploit modularity, e.g., \citep{Liu2017hierarchical}, \citep{Zoph2018learning}, \citep{Negrinho2019modularprogrammable} and \citep{Schrodi2022towards}, ii)~adaptively growing search spaces for refining the architecture, e.g., \citep{Cortes2017adanet}, \citep{Elsken2018efficient} and \citep{Schiessler2021neural}, and iii)~differentiable architecture search, e.g., \citep{Liu2018darts} \citep{Li2020geometry}.
Sparsity promoting training techniques like pruning typically address dense, fully-connected layers and, therefore, are expected to provide little use in the current architecture. Similar reasoning applies to generic regularizers like $\mathcal{L}_2$ regularization. On the other hand, tailored regularizers such as physics-informed losses or the regularizer we proposed in \cite{Guoyue2021metalearning} can prove useful to promote specific algorithmic properties.

More traditional MINLP solvers like those used for superstructure optimization of process systems \citep{Grossmann2002review,Burre2022comparison} exhibit certain advantages over NAS methods.
They explicitly deal with integer variables allowing sophisticated use of discrete choices e.g., for mutually exclusive architecture choices.
Moreover, MINLPs can be solved deterministically to guarantee finding a global solution, e.g., by a branch-and-bound algorithm \citep{Belotti2013mixed}.
Global solution of MINLPs involving the superstructure is challenging, since it requires neural network training subproblems to be solved globally. 
With future advances in computational hardware and algorithms this may become a viable approach.
However, substantial effort is needed to utilize such MINLP solvers for the training tasks considered herein.

\subsection{Implicit layers}\label{sec:oimplicit}

An orthogonal approach for increasing the scope of the superstructure and capitalizing on its modularity is to endow only a subset of its modules or layers with trainable variables.  
The remaining modules that are not subject to meta-optimization can be implemented by so called \textit{implicit layers} that implement their functionality, see, e.g., \citep{Rajeswaran2019iMAML} and \citep{Lorraine2020optimizing}. 
In future work, we plan to emulate the residual minimization of Krylov solvers by substituting the output layer $\FN_n$ of the superstructure with differentiable convex optimization layers \citep{Amos2017differentiable, Agrawal2019differentiable}. Then, only the optimal subspace generation, i.e., the parameters of the $\FN_j$ modules, is left to learn, or the subspace generation is optimized with respect to consecutive minimization being performed in that subspace, respectively.

\subsection{Dynamical systems perspective}\label{sec:oCTalgos}

A joint perspective on neural networks and dynamical systems has emerged recently, e.g., \citep{Weinan2017proposal}, \citep{Haber2017stable}, and \citep{Chang2017multi}. 
Similarly, a connection between dynamical systems and continuous-time limits of iterative algorithms has been discussed in literature \citep{Stuart1998dynamical,Chu2008linear,Dietrich2020koopman}, especially for convex optimization \citep{Su2014Differential,Krichene2015accelerated,Wibisono2016variational}.
Researchers have applied numerical integration schemes to these continuous forms to recover discretized algorithms \citep{Scieur2017integration,Betancourt2018symplectic,Zhang2018directRK}. 
Conversely, the underlying continuous-time dynamics of discrete algorithms encoded by the proposed \R2N2 can be identified based on the iterates they produce, e.g., by their associated Koopman operators \citep{Dietrich2020koopman}. 
These Koopman operators can then be analyzed to compare various algorithms and, even, to identify conjugacies between them \citep{Redman2022algorithmic}.

%% file: sections_rev/acknowledgement.tex
\section*{Declaration of Competing Interest}
We have no conflict of interest.

\section*{Acknowledgements}
\label{sec:acknowledgements}

DTD, AM and MD received funding from the Helmholtz Association of German Research Centres and performed this work as part of the Helmholtz School for Data Science in Life, Earth and Energy (HDS-LEE).
YG and QL are supported by the National Research Foundation, Singapore, under the NRF fellowship (project No. NRF-NRFF13-2021-0005).
FD received funding from the Deutsche Forschungsgemeinschaft (DFG, German Research Foundation) –- 468830823.
The work of IGK is partially supported by the US Department of Energy and the US Air Force Office of Scientific Research. 

%% file: sections_rev/Appendix_A.tex
\section*{SM1. Forward-differencing in the \R2N2 superstructure}\label{app:fwddiff}

\subsection*{SM1.1. Black-box function evaluations}\label{sec:mbbf}

In order to promote discovery of time-proven algorithms, the superstructure can induce some prior structure. For instance, the \mbox{RK-NN} developed in our previous work \citep{Guoyue2021metalearning} essentially sets
\begin{equation*}
    \Vv_j=\Ff(\Vx_k+h \sum_{l=0}^{j-1} a_{j,l} \Vv_l)
\end{equation*}
to promote RK methods with lower-triangular Butcher tableaus.
In the \R2N2 superstructure, this corresponds to Equation~(3.1) with $\Vtheta_j \vcentcolon = \left(a_{j,0},\ldots,a_{j,j-1}\right)$ followed by a direct function evaluation, $\Ff(\Vx'_j)$.  

\begin{figure}[htb]
    \centering
    \includegraphics[scale = 0.4]{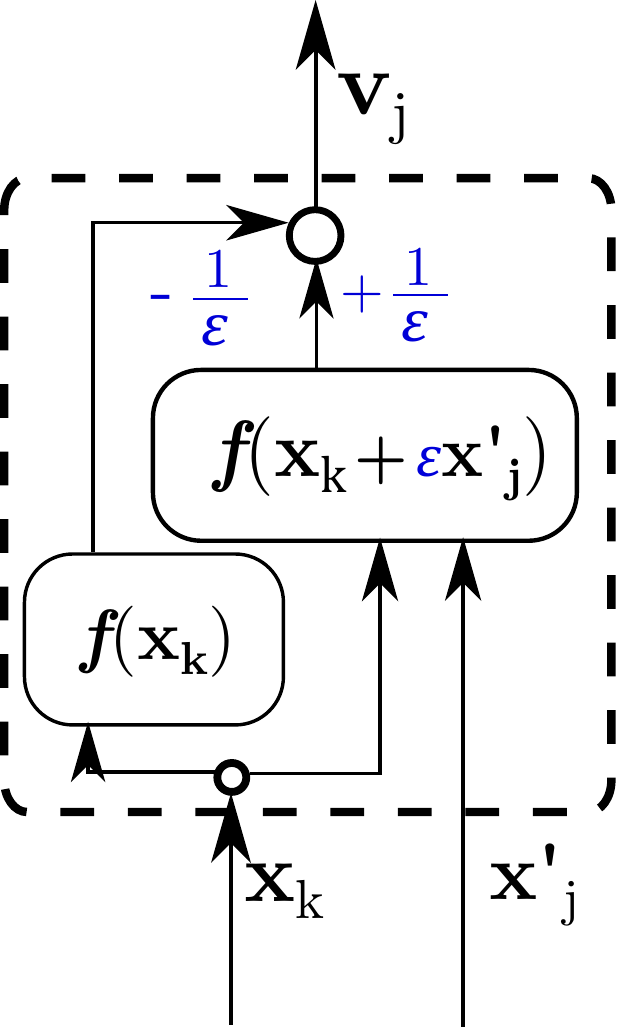}
    \caption{The dashed cell indicates forward-differencing for estimation of directional derivatives with a small $\epsilon$ in the superstructure. The directional derivative at $\Vx_k$ requires reusing the first layer output $\Ff(\Vx_k)$ to explicitly construct finite differences between the outputs of the two function evaluations.}
    \label{fig:black-box_elements}
\end{figure}

We can also hard-wire some of the operations of the superstructure to add yet another implicit bias.
For example, we can encode forward-differencing, i.e.,
\begin{equation}\label{eqn:fwd-diff}
    \Vv_j = \frac{1}{\epsilon}\left(\Ff(\Vx_k + \epsilon \Vx'_j) - \Ff(\Vx_k)\right),
\end{equation}
inside the superstructure for estimation of directional derivatives. Equation~\eqref{eqn:fwd-diff} estimates the derivative of $\Ff$ at $\Vx_k$ in the direction of some $\Vx'_j$. That is, the $j$-th layer then always approximates a directional derivative at the current iterate $\Vx_k$, where only the direction is determined by the trainable weights $\Vtheta_j$ in $\FN_j$ according to Equation~(3.1). 
The corresponding pass through a layer of the superstructure is sketched in \Cref{fig:black-box_elements}. 
While Equation~\eqref{eqn:fwd-diff} and \Cref{fig:black-box_elements} appear to deviate from Equation~(2.7b) and Figure~2, respectively, we next illustrate that forward-differencing is contained as a special case in the superstructure.

\subsection*{SM1.2. Forward-differencing can be expressed by regular function evaluations in the \R2N2 superstructure}\label{app:morphism}

\begin{figure*}[htb] 
    \centering
    \includegraphics[scale=0.35]{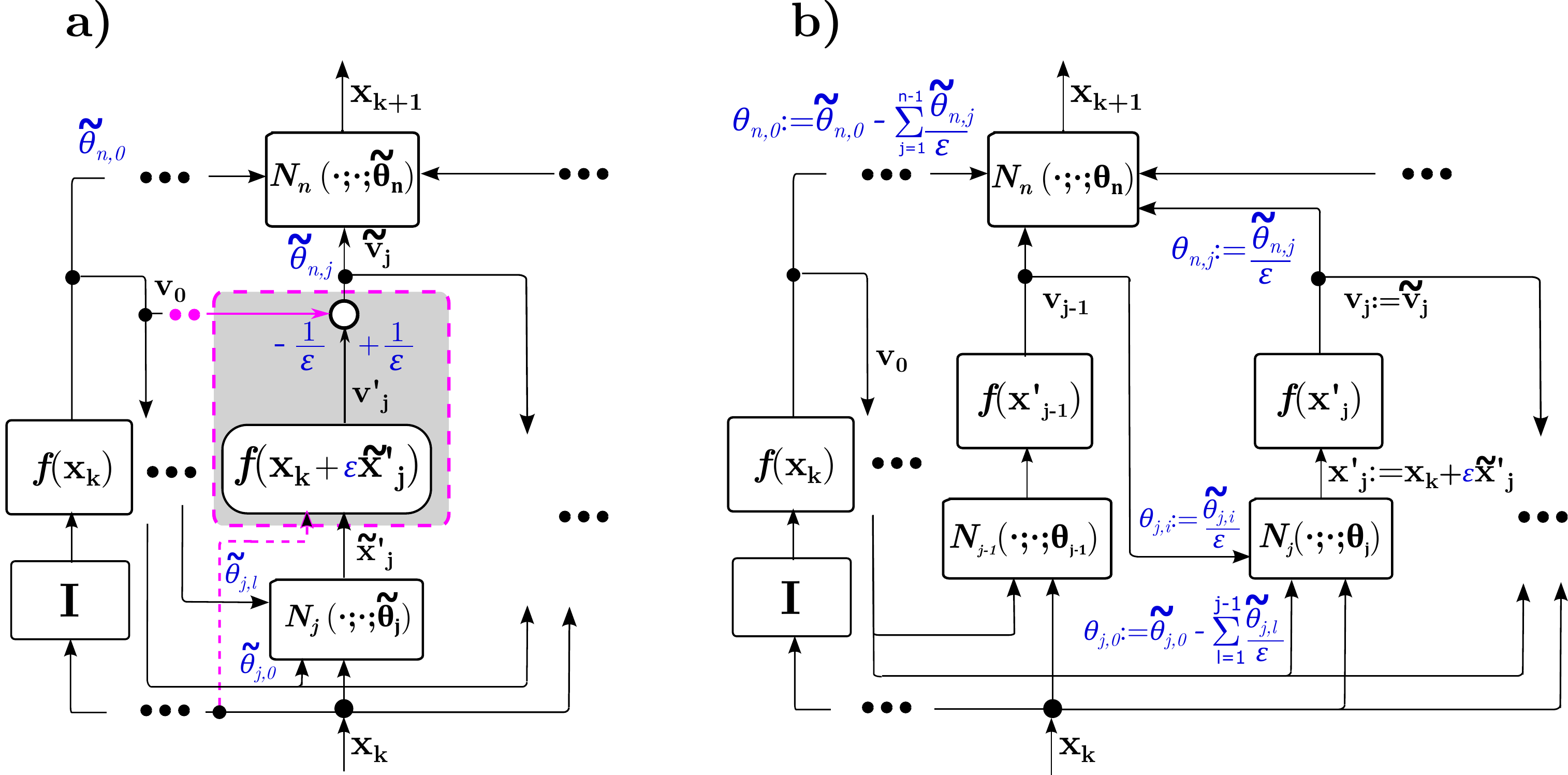}
    \caption{Topological equivalence between one forward-differencing layer a) and a regular function evaluation layer b). 
    The gray box with dashed pink surrounding encapsulates the encoding of a forward differencing. 
    Black symbols are input and output quantities inside the superstructure. The quantities belonging to the forward-differencing encoding are marked by a tilde. 
    Note that two additional inputs are needed for the forward-differencing layer. First, $\Vx_k$ is directly passed to the function evaluation such that $\Vx_k + \epsilon \tilde{\Vx}'$ acts as the function argument where the linear module $\FN_j\left(\tilde{\Vtheta}_j\right)$ outputs an \textit{intermediate} $\tilde{\Vx}'$. Second, $\Ff(\Vx_k) = \Vv_0$ is subtracted from that $\tilde{\Vx}'$ and a division by $\epsilon$ yields the final output $\tilde{\Vv}_j$ of the forward-differencing layer.
    Scalar weights are shown as blue symbols and are related to the trainable parameters.
    In b), on the other hand, $\FN_j\left(\Vtheta_j\right)$ must directly return $\Vx'$ such that the layer output $\Vv_j$ equals $\tilde{\Vv}_j$ from a). 
    Additionally, in b) the transformations from $\tilde{\Vtheta}$ to $\Vtheta$ are shown that result in equivalent output $\Vx_{k+1}$.}
    \label{fig:morphism_fwddiff}
\end{figure*}

\Cref{fig:morphism_fwddiff}~a) illustrates the hard-wiring of forward-differencing operations (Equation~\eqref{eqn:fwd-diff}) in the layers of the \R2N2 superstructure, such that the layer output estimates a directional derivative at $\Vx_k$.
The layer architecture shown in \Cref{fig:morphism_fwddiff}~a) results from a specific parametrization of the general \R2N2 superstructure, see \Cref{fig:morphism_fwddiff}~b).
If the optimal \R2N2 parameters for the forward-differencing configuration are given by $\tilde{\Vtheta}$, there exists a corresponding set of parameters $\Vtheta$ for the \R2N2 structure using regular function calls, which follows directly from $\tilde{\Vtheta}$.
For example, the weighting of the first function evaluation, $\Vv_0$ is transformed by
\begin{equation}\label{eqn:theta0_transform}
    \theta_{j,0} \vcentcolon = \tilde{\theta}_{j,0} - \sum_{l=1}^{j-1} \frac{\tilde{\theta}_{j,l}}{\epsilon},
\end{equation}
where $\epsilon$ corresponds to the value used for calculating finite differences in the finite-differencing module, e.g., $\epsilon = 10^{-8}$.
Moreover, the forward-differencing module in \Cref{fig:morphism_fwddiff}~a) can be thought of as having a fixed bypass with weight of $1$ from the current iterate $\Vx_k$ since it computes a function of $\Vx_k + \epsilon \tilde{\Vx}'$, where $\tilde{\Vx}'$ is the output of $\FN_j\left(\tilde{\Vtheta}_j\right)$. 
To have identical layer outputs $\tilde{\Vv}_j$ and $\Vv_j$, the module $\FN_j\left(\Vtheta_j\right)$ in \Cref{fig:morphism_fwddiff}~b) must directly return $\Vx' \vcentcolon = \Vx_k + \epsilon \tilde{\Vx}'$. 
Note that the subtracted summation over preceding layer indices in Equation~\eqref{eqn:theta0_transform} is due to the construction in \Cref{fig:morphism_fwddiff}~a) and Equation~\eqref{eqn:fwd-diff} that anchors all finite differences at $\Vx_k$. In theory, $\Vtheta_j$ could also express finite differences evaluated at arbitrary positions in the image of $\FN_j$. 
However, the output layer weights $\Vtheta_n$ must change in a complementary fashion in this case, such that the overall inputs to $\FN_n$ are still computed by a forward difference as demanded by Equation~\eqref{eqn:fwd-diff}.

Hard-wiring an operation like forward-differencing may be useful to guide the training procedure. To achieve forward-differencing in the limit, certain trainable parameters must tend to zero. 
\cite{Choi2019empirical} have analyzed a similar problem in the hyperparameter spaces of optimizers and noted a difficulty in learning parameters in different orders of magnitude.
Hence, fixing a routine like forward-differencing \textit{a priori} may be beneficial.

\subsection*{SM1.3. Implementation of forward-differencing modules}\label{app:pytorch_autograd_f}

To fix forward-differencing (Equation~\eqref{eqn:fwd-diff}) in the \R2N2 superstructure also requires a backward pass. We wrapped Equation~\eqref{eqn:fwd-diff} as a PyTorch \textit{Autograd} function and manually implemented the backward passes. 
Backpropagating through the regular function evaluation, $\Ff$, is straightforward:
\begin{equation}\label{eqn:f_bp}
    \frac{\partial \Vv_j}{\partial \Vx'_j}(\Vx'_j) = \frac{\partial \Ff}{\partial \Vx}(\Vx'_j).
\end{equation}
Accordingly, for the backward pass through forward-differencing, we require the derivatives for both of its terms:
\begin{equation}\label{eqn:fwd-diff_bp}
  \begin{aligned}
  \frac{\partial \Vv_j}{\partial \Vx_k}(\Vx'_k,\Vx'_j) = \frac{1}{\epsilon}\left(\frac{\partial \Ff}{\partial \Vx}(\Vx_k + \epsilon \Vx'_j) - \frac{\partial \Ff}{\partial \Vx}(\Vx_k)\right),  \\
  \frac{\partial \Vv_j}{\partial \Vx'_j}(\Vx'_k,\Vx'_j) = \frac{\partial \Ff}{\partial \Vx}(\Vx'_j)
  \end{aligned}
\end{equation}
Note that the derivative with respect to $\Vx_k$ only impacts the weight update for $k > 1$, i.e., when computing at least two iterations with the \R2N2 during training. $\Vx_0$ itself does not depend on trainable parameters.
An alternative to the implementation described herein is to express the forward functions only in terms of PyTorch built-in operations or using other libraries that support automatic differentiation.

%% file: sections_rev/Appendix_B.tex
\section*{SM2. Detailed description of the generated test problems}\label{app:experimental_details}

\subsection*{SM2.1. Linear equation test problems}\label{app:linsolve_details}

In this section, we describe the detailed generation of problem data for linear equation solving problems used in this study (paper and supplementary material). 
We consider linear problems of the form 
\begin{equation*}\label{eqn:lin_solveB}
    \mA\Vx -\Vb = 0,
\end{equation*}
where $\mA\in \mathbb{R}^{m \times m}$, $\Vx\in \mathbb{R}^{m}$, and $\Vb\in\mathbb{R}^{m}$ and we choose $m=5$ as the problem dimensionality. 
Matrices $\mA$ are generated as $\mA=\tilde{\mA}^T \tilde{\mA} + \mI \Vlambda$, where $\tilde{\mA}$ is a matrix with random entries $a_{ij} \sim \mathcal{N}\left(0,0.1\right)$ and $\Vlambda$ can be used to manipulate the spectrum of $\mA$. Here, we use $\Vlambda = \left( 1, 0.75, 0.5, 0.1, 0.1 \right)^T$.
These choices guarantee that $\mA$ is symmetric positive definite with eigenvalues fairly suited for Krylov methods \citep{Kelley1995iterative}, and, in extension, that Problem~(3.3) has a unique solution. 
Different problem instances are completed by a right-hand side $\Vb_i$, which is sampled by adding uniform noise to a fixed randomly-chosen mean $\tilde{\Vb}$, i.e., $\Vb_i = \tilde{\Vb}+\Vb'_i$, where $\tilde{\Vb} \sim \mathcal{N}\left(0,5 \cdot \mI_m\right)$ and $\Vb' \sim \mathcal{U}\left(-\mathbbm{1}^m, \mathbbm{1}^m  \right)$. 
The initial input to the \R2N2 (and to GMRES) is fixed as $\Vx_0 = \mathbf{0}$.  
The matrices $\mA_1$--$\mA_3$ that are used for the training set are generated according to the procedure described above.

Further, we have have used the following matrices for additional experiments in Section~SM3.1.
$\mA_4$--$\mA_7$ are defined, using $\Delta\Vlambda = \left( 0.5, 0.4, 0.3, 0.2, 0.1 \right)^T$, as: 
\begin{subequations}
\begin{equation*}
    \mA_4 = \mA_1 - diag(\Delta\Vlambda),
\end{equation*}
\begin{equation*}
    \mA_5 = \mA_1 + 0.5\cdot diag(\Delta\Vlambda),
\end{equation*}
\begin{equation*}
    \mA_6 = \mA_1 + 1.3\cdot diag(\Delta\Vlambda),
\end{equation*}
\begin{equation*}
    \mA_7 = \mA_1 + 2.5\cdot diag(\Delta\Vlambda).
\end{equation*}
\end{subequations}
$\mA_8$--$\mA_{11}$ are randomly generated symmetric matrices. Each of these matrices has a random scalar, $\sigma_l \sim \mathcal{U}\left(0,5\right)$, for the uniform range and then entries $\tilde{a}_{ij} \sim \mathcal{U}\left(0,\sigma_l\right)$. Finally, $\mA$ is obtained as a symmetric matrix by computing $\tilde{\mA}\tilde{\mA}^T$.

Finally, $\mA_{12}$--$\mA_{19}$ are generated with the same procedure as $\mA_1$ -- $\mA_3$. However, the random entries now use $a_{ij} \sim \mathcal{N}\left(0,\sigma_l \right)$ where 2 matrices are generated for each $\sigma_l \in \{0.3, 0.5, 0.7, 1\}$, i.e., $\mA_{12}$ and $\mA_{13}$ for $\sigma_l=0.3$, $\mA_{14}$ and $\mA_{15}$ for $\sigma_l=0.5$, $\mA_{16}$ and $\mA_{17}$ for $\sigma_l=0.7$, and $\mA_{18}$ and $\mA_{19}$ for $\sigma_l=1.0$.

\newpage
\begin{strip}
\begin{equation*}\label{eqref:def_b1_gmres}
    \tilde{\Vb} = \left( 2.483570, -0.691321, 3.238442, 7.615149, -1.170766\right)^T.
\end{equation*}
\begin{equation*}\label{eqref:def_A1_gmres}
    \mA_1 = \begin{pmatrix}
            1.392232 & 0.152829 & 0.088680 & 0.185377 & 0.156244\\
            0.152829 & 1.070883 & 0.020994 & 0.068940 & 0.141251\\
            0.088680 & 0.020994 & 0.910692 & -0.222769 & 0.060267\\
            0.185377 & 0.068940 & -0.222769 & 0.833275 & 0.058072\\
            0.156244 & 0.141251 & 0.060267 & 0.058072 & 0.735495
          \end{pmatrix},
\end{equation*}
\begin{equation*}\label{eqref:def_A2_gmres}
    \mA_2 = \begin{pmatrix}
            1.122760 & -0.040031 & 0.113992 & 0.068578 & 0.089329\\
           -0.040031 & 0.920757 & 0.085742 & 0.089300 & 0.158474\\
            0.113992 & 0.085742 & 0.896851 & 0.150485 & 0.044783\\
            0.068578 & 0.089300 & 0.150485 & 0.729516 & 0.070168\\
            0.089329 & 0.158474 & 0.044783 & 0.070168 & 1.163038
          \end{pmatrix},
\end{equation*}
\begin{equation*}\label{eqref:def_A3_gmres}
    \mA_3 = \begin{pmatrix}
            1.037577 & 0.120230 & -0.149775 & 0.099841 & 0.169390\\
            0.120230 & 1.095856 & 0.180211 & 0.120029 & 0.133797\\
           -0.149775 & 0.180211 & 0.781548 & 0.241405 & 0.320369\\
            0.099841 & 0.120029 & 0.241405 & 0.877185 & 0.040910\\
            0.169390 & 0.133797 & 0.320369 & 0.040910 & 0.602205
          \end{pmatrix},
\end{equation*}
\begin{equation*}\label{eqref:def_A8_gmres}
    \mA_8 = \begin{pmatrix}
             9.801337 & -4.563474 & 2.196806 & -5.154676 & 5.063176\\
            -4.563474 & 48.751049 & -26.335994 & -3.910831 & 17.380485\\
             2.196806 & -26.335994 & 31.887071 & 1.215492 & -12.532923\\
            -5.154676 & -3.910831 & 1.215492 & 4.0743960 & -5.876128\\
             5.063176 & 17.380485 & -12.532923 & -5.876128 & 25.900849
          \end{pmatrix},
\end{equation*}
\begin{equation*}\label{eqref:def_A9_gmres}
    \mA_9 = \begin{pmatrix}
            19.582102 & -1.721533 & 5.067191 & 20.194875 & 1.561468\\
           -1.721533 & 38.090555 & -11.445662 & -22.832142 & -0.152421\\
            5.067191 & -11.445662 & 17.191893 & 14.784228 & -3.889048\\
            20.194875 & -22.832142 & 14.784228 & 49.221081 & 22.059518\\
            1.561468 & -0.152421 & -3.889048 & 22.059518 & 31.461613
          \end{pmatrix},
\end{equation*}
\begin{equation*}\label{eqref:def_A10_gmres}
    \mA_{10} = \begin{pmatrix}
            1.543741 & -1.708336 & -0.855255 &  1.180115 & -0.606022\\
           -1.708336 &  7.993454 &  1.813288 & -0.855154 & -0.375811\\
           -0.855255 &  1.813288 &  2.131294 & -2.223852 & -0.808170\\
            1.180115 & -0.855154 & -2.223852 &  3.296235 &  1.148258\\
           -0.606022 & -0.375811 & -0.8081702 & 1.148258 &  2.018821
          \end{pmatrix},
\end{equation*}
\begin{equation*}\label{eqref:def_A11_gmres}
    \mA_{11} = \begin{pmatrix}
           0.554750 & 0.192700 & -0.030087 & -0.173792 & 0.078237\\
           0.192700 &  0.134709 & 0.005420 & 0.156018 & -0.081507\\
          -0.030087 & 0.005420 & 0.491319 & -0.087115 & -0.068497\\
          -0.173792 & 0.156018 & -0.087115 & 0.923782 & -0.356224\\
           0.078237 & -0.081507 & -0.068497 & -0.356224 & 0.197102
          \end{pmatrix}.
\end{equation*}
\begin{equation*}\label{eqref:def_A12_gmres}
    \mA_{12} = \begin{pmatrix}
            1.803328 & 0.200759 & -0.355809 & -0.098682 & -0.037251\\
           0.200759 & 1.243347 & 0.088843 & 0.263899 & 0.195536\\
           -0.355809 & 0.088843 & 1.495596 & 0.093483 & 0.383077\\
           -0.098682 & 0.263899 & 0.093483 & 1.295673 & 0.091526\\
           -0.037251 & 0.195536 & 0.383077 & 0.091526 & 1.171966
          \end{pmatrix},
\end{equation*}
\begin{equation*}\label{eqref:def_A13_gmres}
    \mA_{13} = \begin{pmatrix}
            1.373797 & 0.029822 & 0.291240 & -0.06804 & -0.122712\\
           0.029822 & 1.352286 & 0.213403 & 0.259224 & 0.113595\\
           0.291240 & 0.213403 & 1.145153 & 0.260138 & -0.256945\\
           -0.068040 & 0.259224 & 0.260138 & 1.044292 & 0.023357\\
           -0.122712 & 0.113595 & -0.256945 & 0.023357 & 1.493027
          \end{pmatrix},
\end{equation*}
\begin{equation*}\label{eqref:def_A14_gmres}
    \mA_{14} = \begin{pmatrix}
             1.875641 & 0.369074 & -0.254450 & 0.011282 & 0.086120\\
           0.369074 & 1.438546 & 0.165303 & 0.330450 & 0.326974\\
           -0.254450 & 0.165303 & 1.578616 & 0.135095 & 0.435910\\
           0.011282 & 0.330450 & 0.135095 & 1.407443 & 0.175663\\
           0.086120 & 0.326974 & 0.435910 & 0.175663 & 1.302648
          \end{pmatrix},
\end{equation*}
\begin{equation*}\label{eqref:def_A15_gmres}
    \mA_{15} = \begin{pmatrix}
           1.496302 & 0.069012 & 0.466847 & -0.023807 & -0.07450\\
           0.069012 & 1.356091 & 0.304412 & 0.368689 & 0.278316\\
           0.466847 & 0.304412 & 1.273936 & 0.359224 & -0.193665\\
           -0.023807 & 0.368689 & 0.359224 & 1.096486 & 0.099104\\
           -0.0745018 & 0.278316 & -0.193665 & 0.099104 & 1.661732
          \end{pmatrix}.
\end{equation*}
\begin{equation*}\label{eqref:def_A16_gmres}
    \mA_{16} = \begin{pmatrix}
            1.947954 & 0.537389 & -0.153091 & 0.121248 & 0.209492\\
            0.537389 & 1.633745 & 0.241763 & 0.397001 & 0.458412\\
           -0.153091 & 0.241763 & 1.661637 & 0.176707 & 0.488742\\
           0.121248 & 0.397001 & 0.176707 & 1.519214 & 0.259799\\
           0.209492 & 0.458412 & 0.488742 & 0.259799 & 1.433331
          \end{pmatrix},
\end{equation*}
\begin{equation*}\label{eqref:def_A17_gmres}
    \mA_{17} = \begin{pmatrix}
           1.618807 & 0.108202 & 0.642453 & 0.020426 & -0.026291\\
           0.108202 & 1.359896 & 0.395421 & 0.478153 & 0.443036\\
           0.642453 & 0.395421 & 1.402719 & 0.458310 & -0.130384\\
           0.020426 & 0.478153 & 0.458310 & 1.148681 & 0.174850\\
           -0.026291 & 0.443036 & -0.130384 & 0.174850 & 1.830438
          \end{pmatrix}.
\end{equation*}
\begin{equation*}\label{eqref:def_A18_gmres}
    \mA_{18} = \begin{pmatrix}
            2.056424 & 0.789862 & -0.001053 &  0.286197 & 0.394551\\
            0.789862 & 1.926544 & 0.356453 & 0.496827 & 0.655569\\
           -0.001053 & 0.356453 & 1.786167 & 0.239125 & 0.567990\\
            0.286197 & 0.496827 & 0.239125 & 1.686871 & 0.386004\\
           0.394551 & 0.655569 & 0.567990 & 0.386004 & 1.629354
          \end{pmatrix},
\end{equation*}
\begin{equation*}\label{eqref:def_A19_gmres}
    \mA_{19} = \begin{pmatrix}
            1.802565 & 0.166987 & 0.905863 & 0.086776 & 0.046025\\
           0.166987 & 1.365604 & 0.531934 & 0.642350 & 0.690117\\
           0.905863 & 0.531934 & 1.595893 & 0.606940 & -0.035464\\
           0.086776 & 0.642350 & 0.606940 & 1.226972 & 0.288470\\
           0.046025 & 0.690117 & -0.035464 & 0.288470 & 2.083496
          \end{pmatrix}.
\end{equation*}
\end{strip}

\subsection*{SM2.2. Nonlinear equation test problems}\label{app:fsolve_details}

As a case study for nonlinear problems we use a discretization of Chandrasekhar's $H$-function in the conservative case, which has also served as a test problem for nonlinear methods in \cite{Kelley2003solving}. We give the derivation by \cite{Kelley2003solving} in the following. The $H$-function reads:
\begin{equation}\label{eqn:chand_H_cont}
   F(H)(\mu) = H(\mu) - \left( 1 - \frac{c}{2} \int_0^1 \frac{\mu H(\nu)d\nu}{\mu+\nu} \right)^{-1} = 0.
\end{equation}
A discretization of \eqref{eqn:chand_H_cont} uses the composite midpoint rule, i.e., \citep{Kelley2003solving}
\begin{equation}\label{eqn:midpoint_rule}
    \int_0^1 f(\mu)d\mu =  \frac{1}{N} \sum_{\iota=1}^m f(\mu_\iota),
\end{equation}
to approximate the integral over $\mathcal{\iota}=1,\ldots,m$ discretization points (and $j=1,\ldots,m$ likewise as a secondary index for the discretization points). 
With setting $\mu_\iota = \frac{(\iota-\frac{1}{2})}{m}$ and $x_j = H(\mu_j)$, the resulting discretized form for each component $j$ of $F$ can be written as \citep{Kelley2003solving}
\begin{equation}\label{eqn:chand_H_discA}
    F(\mathbf{x})_j = x_j - \left( 1- \frac{c}{2m} \sum_{\iota=1}^m \frac{\mu_j x_\iota}{\mu_j + \mu_\iota}\right)^{-1}.
\end{equation}
By defining a matrix $\mA_c \in \mathbb{R}^{m \times m}$ with entries that depend on $m$ and $c$, i.e.,
\begin{equation}\label{eqn:A_c}
    A_{c,j\iota} = \frac{c \mu_j}{2m \left(\mu_j+\mu_\iota\right)},
\end{equation}
we can simplify the above discretization to \citep{Kelley2003solving}
\begin{equation}\label{eqn:chand_H_discB}
    \left(F(\Vx)\right)_j = \Vx_j - \left(1 - (\mA_c\Vx)_j \right)^{-1}.
\end{equation}

In one of Kelley's examples, $m=10$ and $c=0.9$ were used. We select these values as a basis for the experiments in our work. We fix $m=10$ discretization points but we vary the parameter $c$ of Equation \eqref{eqn:chand_H_discB} in $\left\{0.875,0.905,0.935\right\}$ in order to generate a set of training problems. $c \in \left\{ 0.85, 0.95\right\}$ are used for extrapolation.
We further sample initial values from a normal distribution with mean $\tilde{\Vx}_0=\mathbbm{1}^{(m)}$, where $\mathbbm{1}^{(m)}$ is an $m$-dimensional vector of ones, and variance of $\Vsigma=0.2\cdot\mathbbm{1}^{(m)}$, such that the entries of $\tilde{\Vx}_0$ are uncorrelated.

%% file: sections_rev/Appendix_C.tex
\section*{SM3. Additional numerical experiments}\label{app:r_app}

In this section, we present additional numerical experiments. These include a more detailed examination of the extrapolation studies done in Sections~4.1 and 4.2 (Section~SM3.1), a comparison of the \R2N2 to vanilla neural networks (Section~SM3.2), and an ablation study for an \R2N2 with iteration-dependent parameters (Section~SM3.3).

\subsection*{SM3.1. Detailed generalization and extrapolation experiments for linear equation solving}\label{app:rapp_extrapol}

\begin{figure*}[htb]
    \centering
    \includegraphics[scale=1.0]{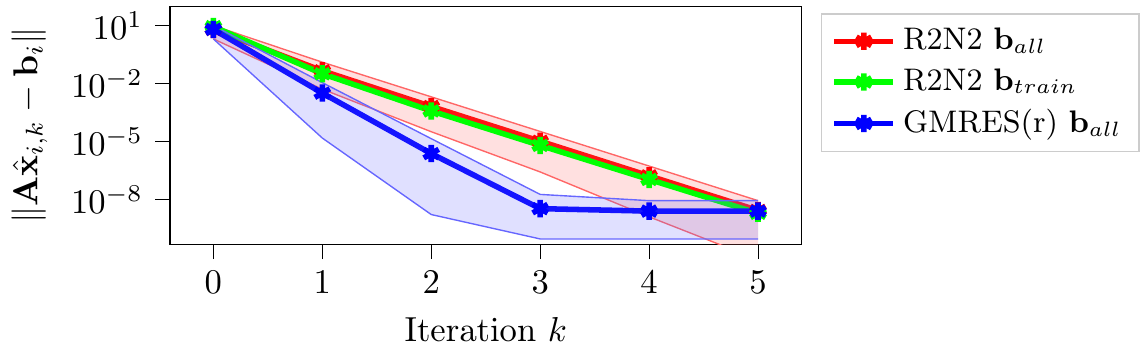}
    \caption{Convergence of the \R2N2 for linear problems with $\mA_1$ and arbitrary right-hand sides from $[-5,5]^m$ (denoted by $\Vb_{all}$). The solid line indicates the mean, whereas the area shows the spread between minimum and maximum performance.
    The \R2N2 (mean) performance on the training distribution is given by the green line (denoted by $\Vb_{train}$), and is only slightly better than that for arbitrary RHS (red line).}
    \label{fig:lin_conv_all_b}
\end{figure*}

We analyze the extrapolation results of the linear \R2N2 shown in Figure~5 in more detail.
First, we consider extrapolation in the space of RHS for the fixed matrix $\mA_1$, i.e., problems $\left(\mA_1, \Vb_i \right)$ with $\Vb_i$ from a range not used in training. 
Given that $\Vb_i$ were constructed by adding noise around a specific mean $\tilde{\Vb}$, straightforward experiments are to increase the noise or to generate samples around a different base $\tilde{\Vb}'$.
However, as we already observed empirically (and will further show in Section~SM4), the \R2N2 is globally convergent towards solutions for all $\Vb$ given $\mA_1$. 
We thus only analyze the performance for a set $\{\Vb'\}_i$ of $500$ samples, where $\tilde{\Vb} = \mathbf{0}$ and the additive noise is $\Vb' \sim \mathcal{U}\left(-5\cdot\mathbbm{1}^m, 5\cdot\mathbbm{1}^m  \right)$. We choose this interval, since the initial training set had $\Vert\tilde{\Vb}\Vert = 5$, and we want $\Vr_0$ to be of similar magnitude for all samples.
The results, shown in \Cref{fig:lin_conv_all_b}, show that the \R2N2 converges for all of these problem instances. The spread between the minimum and the maximum at each iteration is tolerable too, with a difference in convergence rate of less than a factor of 2. Further, the mean of the training samples, plotted in orange, is only slightly better than that of the test sample of arbitrary RHS.

\begin{figure*}[htb] 
    \centering
    \includegraphics[scale=1.0]{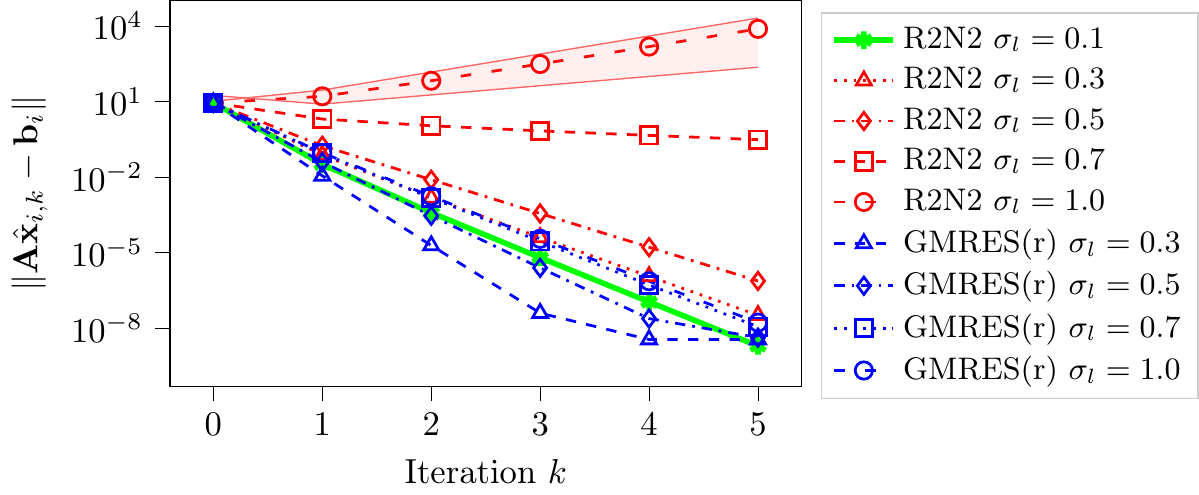}
    \caption{Convergence of the \R2N2 for linear problems with increased levels of noise in the base entries of $\mA$, cf.~Section~SM2.1. 
    Performance on the original training set is given by the green line. Compared to that, the noise level is gradually increased to a tenfold. The convergence of the \R2N2, indicated by the red lines, decreases with increased noise until, eventually for noise level $\sigma_l=1$, the \R2N2 iterations lead to divergence on the test problems.
    We also plot the range between the minimum and maximum samples for $\sigma_l=1$, highlighting that the divergence in mean is not caused by outliers.}
    \label{fig:lin_conv_morenoiseA}
\end{figure*}

Next, we analyze application of the \R2N2 to problems, where matrix $\mA$ has a higher base noise than $\mA_1$--$\mA_3$. That is, we deploy the \R2N2 on a test set generated from the original set of $\Vb_i$ and a pair of matrices with base entries $a_{ij} \sim \mathcal{N}\left(0,\sigma_l\right)$ for each $\sigma_l \in \{0.3, 0.5, 0.7, 1\}$, i.e., we generate a total of 8 new matrices, listed as $\mA_{12}$--$\mA_{19}$ in Section~SM2.1.
\Cref{fig:lin_conv_morenoiseA} shows the performance of the \R2N2 (compared to GMRES(r)). The green line with $\sigma_l=0.1$ corresponds to the training settings. For smaller levels of noise, the performance of the \R2N2 seems to decrease proportionally. However, towards $\sigma_l=0.7$, there is a larger deterioration, where the \R2N2 is still convergent but with a very slow convergence rate.
Finally, at $\sigma_l=1$, the \R2N2 diverges.

\begin{figure*}[htb] 
    \centering
    \includegraphics[scale=1.0]{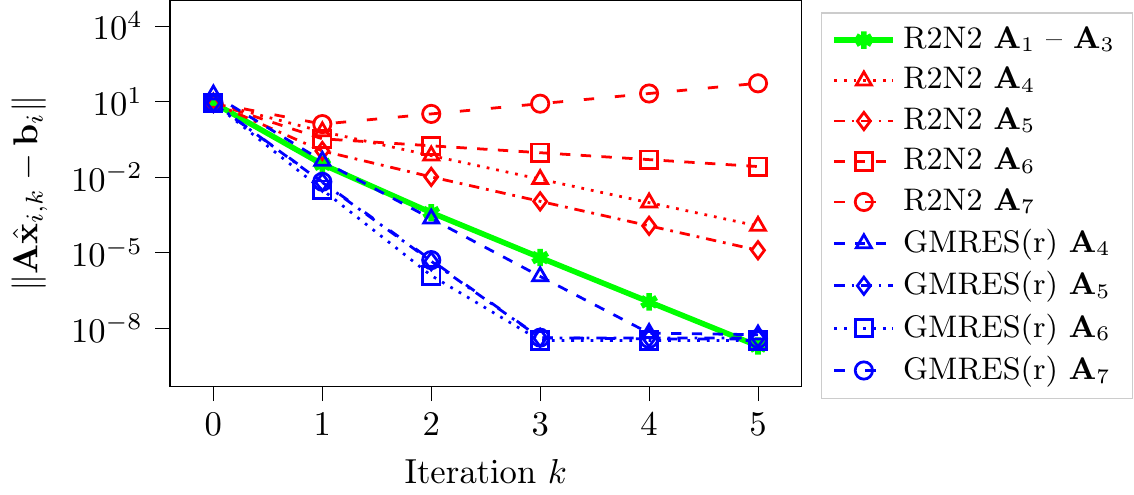}
    \caption{Convergence of the \R2N2 for linear problems when the diagonal entries of $\mA$ are decreased or increased respectively.
    Performance on the original training set is given by the green line. 
    For changes in either direction, the \R2N2 performance (red lines) worsens. This underscores that the \R2N2 is \textit{adapted} to a certain problem class, from which we move away by the changes to the diagonals.}
    \label{fig:lin_conv_extrapol_lambda}
\end{figure*}

The test problems we use in this paper overlay the noisy base entries of $\mA$ with a spectrum $\Vlambda$. 
We now examine how the convergence behavior of the \R2N2 changes when $\Vlambda$ is amplified or attenuated, respectively. 
For this study, we use matrices $\mA_4$--$\mA_7$ combined with the original sample of RHSs $\left\{\Vb_i\right\}$ to form a test set for the \R2N2.
In \Cref{fig:lin_conv_extrapol_lambda}, we analyze the resulting performance of the \R2N2 on that test set.
Changes to the overlaid spectrum in either direction reduce the performance of the \R2N2. For $\mA_7$, we even find divergent behavior, see Section~SM4 for an explanation.

\begin{figure*}[htb] 
    \centering
    \includegraphics[scale=1.0]{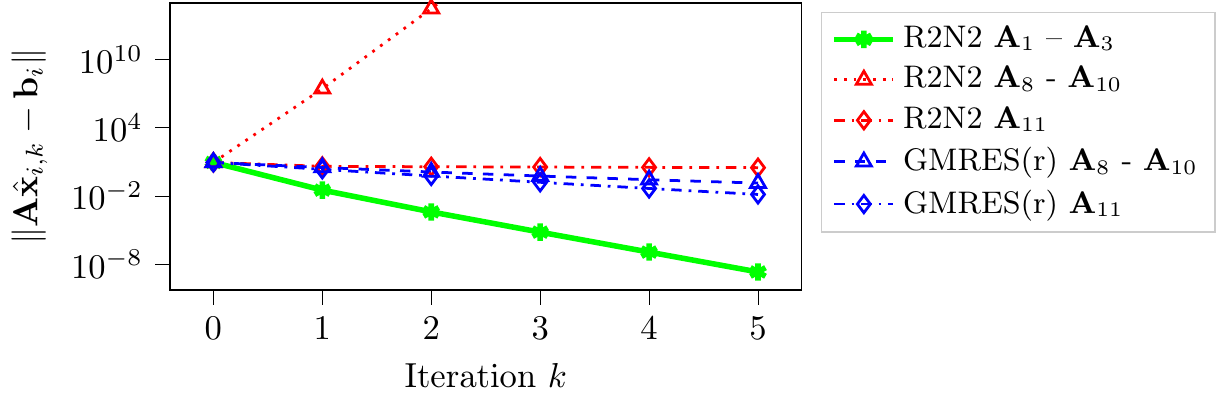}
    \caption{Convergence of the \R2N2 (shown by red lines) for linear problems formed from random symmetric matrices $\mA_8$--$\mA_{11}$.
    Performance on the original training set is given by the green line. 
    In general, the \R2N2 does not converge for such random problems that are not related to the training set. However, as $\mA_{11}$ proves, coincidental convergence is possible.}
    \label{fig:lin_conv_fullyRandomA}
\end{figure*}

Finally, we generate four random, symmetric matrices $\mA_8$ -- $\mA_{11}$. For each matrix we have drawn a random scalar $\sigma_l \sim \mathcal{U}\left(0,5\right)$ and entries $\tilde{a}_{ij} \sim \mathcal{U}\left(0,\sigma_l\right)$ and then computed $\tilde{\mA}\tilde{\mA}^T$ to obtain an symmetric matrix. We again use the original sample of RHSs $\left\{\Vb_i\right\}$. 
The performance of the \R2N2 on these random symmetric problems is plotted in \Cref{fig:lin_conv_fullyRandomA}.
The first three of these matrices lead to problem instances, on which the \R2N2 diverges. 
The final matrix, $\mA_{11}$, happened to have properties such that the \R2N2 is actually able to converge to the solutions of the associated test set.
We explain that this is a (lucky) property of $\mA_{11}$ in Section~SM4.

\subsection*{SM3.2. Comparison to vanilla neural networks}\label{app:rapp_NN}

In this subsection, we train a vanilla NN on the same problem data as the \R2N2. The NN has two hidden layers with $20$ neurons each and uses $\tanh$ as the activation function for the hidden layers and linear for the output layer.  

\begin{figure*}[htb] 
    \centering
    \includegraphics[scale=1.0]{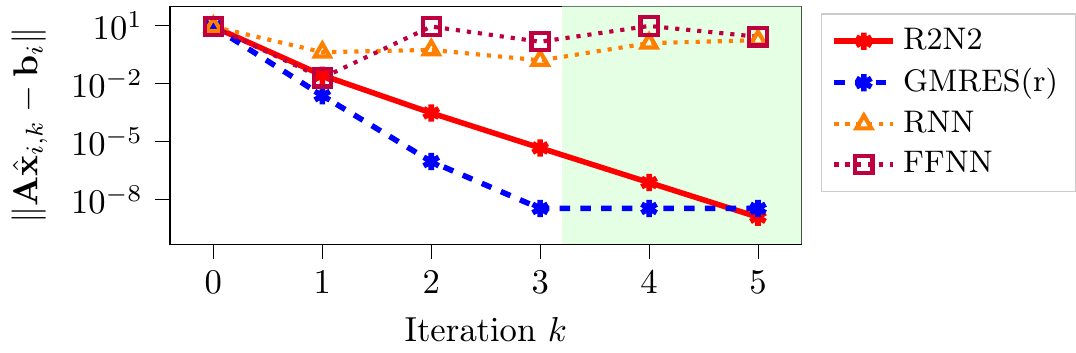}
    \caption{Comparing the convergence behavior of the \R2N2 (red, asterisk) against that of a FFNN (purple, square) and RNN (orange, triangle) trained on the same linear problem data, respectively. 
    The FFNN is trained on a single forward pass, i.e., to minimize the residual at iteration $k=1$.
    The RNN is trained on a weighted loss over iterations $k=\{1,2,3\}$. 
    The green area denotes iterations for which the RNN, \R2N2 (and FFNN) have not been trained.
    GMRES(r) (blue, asterisk) is added as a benchmark.}
    \label{fig:lin_NN_comparison}
\end{figure*}

We examine the linear equation solving problems presented in Section~4.1.
The NN is trained to learn $\left( \mA, \Vb\right) \mapsto \Vx^{\star}$. To this end, the input layer concatenates $\mA$ and $\Vr_0$, where $\Vr_0=-\Vb$, to a $(m^2 + m)$-dimensional, i.e., $30$-dimensional input vector. Owing to this large dimension, the NN has about one order of magnitude more trainable parameters and operations during inference than the \R2N2. For training, we used \textit{Adam} with default settings on $10,000$ epochs. The training data set corresponds to the one used in Section~4.1.
We have trained one NN, the feed-forward NN (\textit{FFNN}), as a direct problem-to-solution mapping, i.e., on a single forward pass.
Further, we trained a second NN, the recurrent (\textit{RNN}), on 3 recurrent iterations using the same loss as for the \R2N2, i.e., Equation~(4.1) with $w_k = 4^k$. Here, the input to the first iteration uses $\left( \mA, \Vr_0\right)$ with $\Vr_0 = -\Vb$. For subsequent iterations, we use the RNN output $\hat{\Vx}_k$ to compute $\Vr_k = \mA\hat{\Vx}_k - \Vb$.
The training of the FFNN has resulted in a final train loss of $1.28e^{-3}$ and test loss of $1.45e^{-3}$, whereas for the RNN we find a train loss of $4.669$ coupled with a test loss of $4.688$.
\Cref{fig:lin_NN_comparison} compares the FFNN and the RNN to the \R2N2 and GMRES on the test set of the training data, i.e., unseen samples of $\Vb_i$, and also for additional recurrences.
The FFNN, which was only trained on one iteration, immediately incurs a drop in performance at the second iteration. The RNN performance, too, decreases when performing additional iterations $k > 3$.
Clearly, neither the FFNN nor the RNN learn a behavior that can match the performance of the \R2N2 despite possessing much more trainable parameters than the latter.

\begin{figure*}[htb] 
\centering
\includegraphics[scale=1.0]{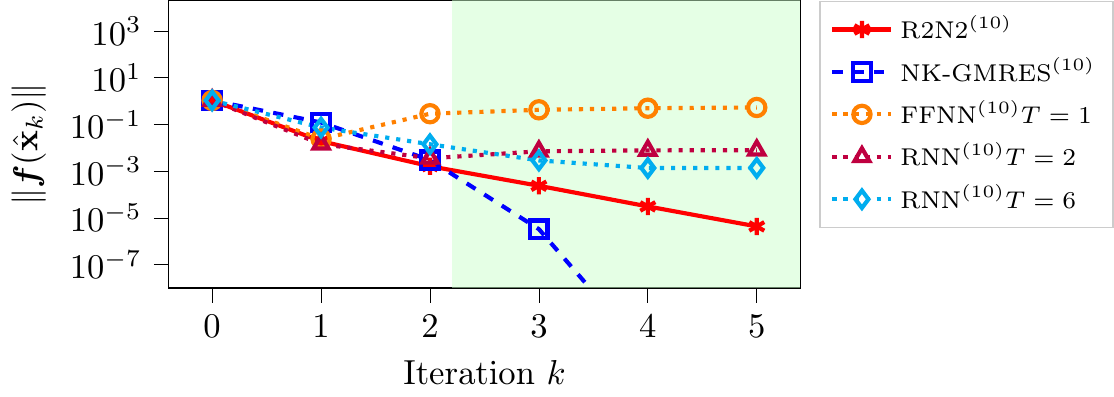}
\caption{Comparing the convergence behavior of the \R2N2 (red, asterisk) against that of various NNs trained on the same nonlinear problem data. 
The FFNN (orange, circles) is trained on a single forward pass, i.e., to minimize the residual at iteration $k=1$.
RNN$_2$ (purple, triangles) is trained trained on a weighted loss over iterations $k=\{1,2\}$, like the \R2N2.
RNN$_6$ (cyan, diamonds) is trained on a weighted loss over iterations $k=\{1,\ldots,6\}$, which allows to see (and operate on) the same number of function evaluations as the \R2N2.
NK-GMRES (blue, asterisk) is added as a benchmark.}
\label{fig:nonlin_NN_comparison}
\end{figure*}

Next, we compare the vanilla NN models to the \R2N2 on the discretized form of Chandrasekhar's $H$-function (Equation~(4.4)). 
As a training set, we use the one generated for the \R2N2 in Section~4.2. However, since the NN is not agnostic to the dimension of inputs $\Vx_0$, we only use the subset where $m=10$ for training and comparison herein.
We provide the NNs with the input $\left( \Vx_k, \Ff(\Vx_k)\right)$ , i.e., we allow for function evaluations to happen externally to the NNs.
We train three different NNs: the \textit{FFNN} is again trained on a single forward pass, the \textit{RNN}$_2$ is trained on two recurrent iterations to match the \R2N2, and the \textit{RNN}$_6$ is trained on 6 consecutive iterations, such that it can learn to operate with the same amount of function evaluations as the \R2N2 was allowed in this case. 
Owing to the comparatively large hidden layers, the NNs have about one or two orders of magnitude more trainable parameters and overall operations in a forward pass than the \R2N2.
\Cref{fig:nonlin_NN_comparison} compares the three NNs to the \R2N2 and NK-GMRES on the test set of the training data, i.e., unseen samples of $\Vx_0$.
Again, all three NNs achieve a partial reduction of the residual in accordance with the training objective, but cannot match the performance of the \R2N2, especially over more than 2 iterations. The trained NNs are neither able to ``predict'' solutions to the test problems more accurately than the \R2N2 or NK-GMRES, nor do they exhibit a behavior indicative of a solver, when applied recurrently.
We therefore omit further studies and conclude that the \R2N2 represents iterative solvers much better than vanilla-architecture neural networks that learn pure input-to-output mappings.

\subsection*{SM3.3. Multiple outer iterations of linear equation solving}\label{app:rapp_restartedR2N2}

\begin{figure*}[htb] 
    \centering
    \includegraphics[scale=1.0]{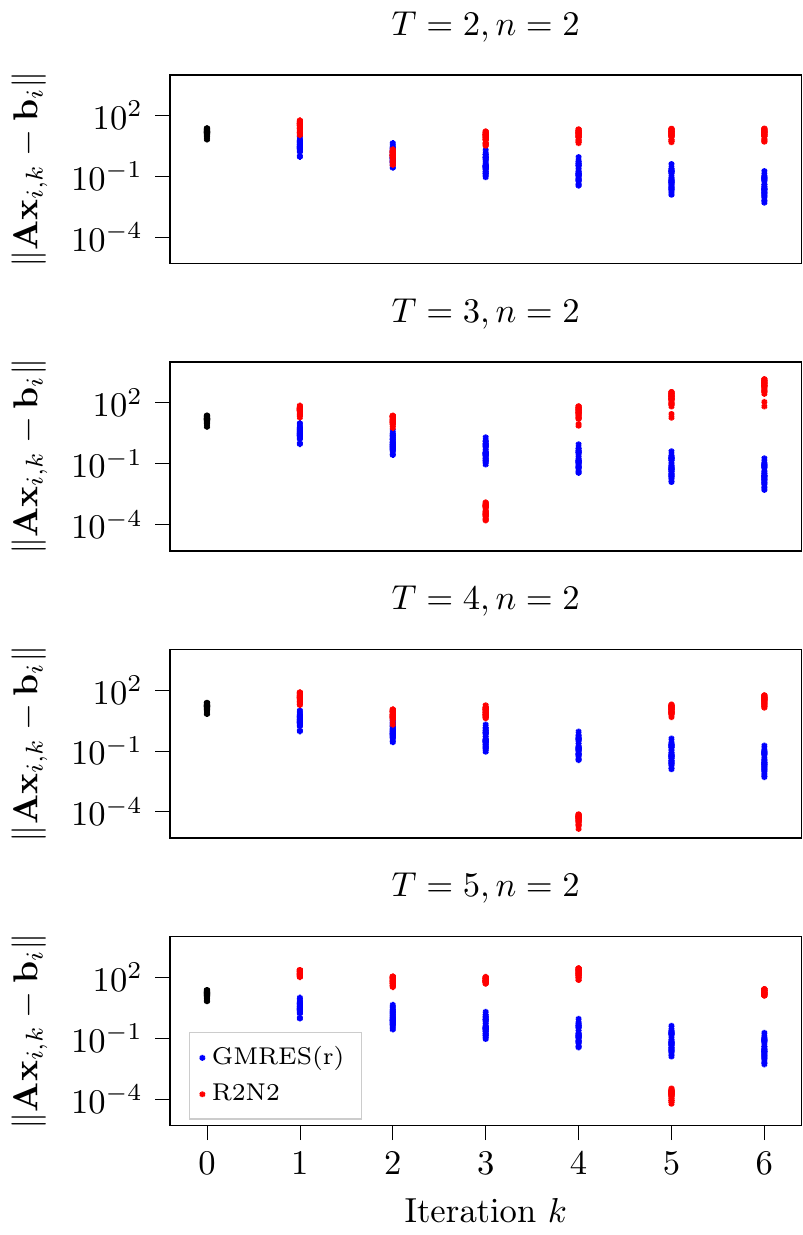}
    \caption{Convergence of the \R2N2 vs GMRES for $6$ nonlinear iterations of solving $\mA\Vx = \Vb_i$ for a random draw of samples $\Vb_i$. The generation of problem data follows the procedure described in (\textbf{SM2}). Both methods use $2$ inner iterations. The different \R2N2 used in the subplots were trained on the loss after $T=\left\{2,3,4,5\right\}$ nonlinear iterations. The parameters in the final module, $\Vtheta_{n,k}$, were relaxed to take varying values across iterations $k\leq T$.}
    \label{fig:lin_multiparam_conv_finalloss}
\end{figure*}

In Section~4.1, we have explored the convergence of the \R2N2 with a fixed $\Vtheta_n$ and a loss term summed over all outer iterations. With that configuration, the \R2N2 was consistently outperformed by GMRES due to GMRES' minimization of the residual in the subspace.
Here, we endow the \R2N2 with varying $\Vtheta_{n,k}$ for iterations $k=1,\ldots,T$ but the training loss only considers the final iterate $\Vx_{i,T}$ of problem instances, i.e.,
\begin{equation*}
    MSE_{f}=  \sum_{i=1}^N \left(\mA\hat{\Vx}_{i,T}-\Vb_i \right)^2,
\end{equation*}
The results of this study are given by Figure~\ref{fig:lin_multiparam_conv_finalloss}. With the additional iteration-dependent parameters, the \R2N2 appears to learn to combine consecutive iterations in order to capture the full space of the residual such that an abrupt step towards the true solution is facilitated in the final iteration. 
This is underscored by the change of behavior between $T=2$ and $T=3$. As soon as the aggregated dimension of the subspaces built during the iterations -- controlled by $n \times T$ -- exceeds the problem dimensionality $m=5$, the \R2N2 is able to achieve a sudden improvement in performance, compared to GMRES. The level of that improvement, however, stagnates for $T > \frac{m}{n}$. 
The remaining residual presumably originates from the variance of the training data, since the \R2N2 cannot adapt to individual problem instances.
Evaluating the \R2N2 for further iterations, which we performed by reapplying the parameters of the final step $\Vtheta_{n,T}$, does not lead to convergence for the steps $k > T$. 
Therefore, by the addition of more trainable parameters, which corresponds to a relaxation of the prior structure of the \R2N2, the \R2N2 behaves more comparable to the NNs analyzed in Section~SM3.2.
That is, improvements over GMRES in the training objective become possible, at the cost of forfeiting the ability to extrapolate or converge to solutions.

%% file: sections_rev/Appendix_D.tex
\section*{SM4. Convergence properties of the \R2N2 on linear problems}\label{app:r_conv}

We analyze the convergence properties of the \R2N2 in more detail.
Results in Section~4.1, Figure~5, have demonstrated that, empirically, the \R2N2 can converge towards the solution of a linear problem for arbitrary right-hand sides $\Vb_i$ in some cases. 
Evidently, the property is then due to the matrix $\mA$ and the coefficients $\Vtheta$ of the \R2N2.

For linear problems, it is straightforward to write how the \R2N2 acts on the (sequence of) residuals $\Vr_k$, given that $\mA$ is a square and invertible matrix.
We write the forward pass as
\begin{equation}\label{eqn:fwd_update}
    \Vx_{k+1} = \Vx_k + \Vzeta \mV,
\end{equation}
where
\begin{equation}\label{eqn:subpsace_sumform}
    \Vzeta \mV = \sum_{j=0}^{n-1} \mA^j \Vr_k \zeta_{j+1},
\end{equation}
Note that the Krylov subspace in the \R2N2 uses coefficients $\Vtheta$ instead of $\Vzeta$ to express the linear combination. $\Vzeta$ do not correspond to the final layer $\Vtheta_n$ in Equation~(3.2).
Instead, the whole of $\Vtheta$ can be transformed into $\Vzeta$. We use the notation with $\Vzeta$ here for more clarity.

Next, we also state the forward pass in terms of the residual inserting Equation~\eqref{eqn:fwd_update}, i.e.,
\begin{equation}\label{eqref:res_update}
    \Vr_{k+1} = \mA \Vx_{k+1} - b = \mA \Vx_k + \mA \Vzeta \mV.
\end{equation}
Further, by inserting for $\Vx_k = \mA^{-1}(\Vr_k + \Vb)$ and Equation~\eqref{eqn:subpsace_sumform}, we obtain
\begin{equation*}
    \Vr_{k+1} = \mA\mA^{-1} \left(\Vr_k + \Vb \right) + \mA\sum_{j=0}^{n-1} \mA^j \Vr_k \zeta_{j+1} - \Vb,
\end{equation*}
which, after canceling $\mA\mA^{-1}$ and $\Vb$, simplifies to
\begin{equation}\label{eqn:r2n2_linoperator}
    \Vr_{k+1} = \left[ \mI+ \sum_{j=1}^n \mA^j  \zeta_{j}\right] \Vr_k = \colon \mathcal{A}^{iter} (\Vr_k),
\end{equation}
where $\mathcal{A}^{iter}$ is the \textit{algorithm operator} encoded by the \R2N2 with parameters $\Vzeta$ (or $\Vtheta$, respectively) for a specific input matrix $\mA$.
If the spectral norm of $\mathcal{A}^{iter}$ (largest singular value) is smaller than~$1$, the \R2N2 will converge for all right-hand sides.
For some $\mA$, and given parameters $\Vtheta$, we compute this norm via singular value decomposition.
For instance, the spectral norm of $\mathcal{A}^{iter}$ computed for matrix $\mA_1$ and the trained parameters $\Vtheta$ of the \R2N2 used for the results in Section~4.1 is $0.0163$, which confirms that the \R2N2 will converge for all RHS for $\mA_1$.
On the other hand, $\mathcal{A}^{iter}$ for matrix $\mA_{6}$ (see \Cref{fig:lin_conv_extrapol_lambda}) has spectral norm $0.5315$, which coincides with the slower convergence observed for $\mA_6$.
For $\mA_7$ we even have spectral norm $2.5397$ and the \R2N2 does indeed diverge on problem instances featuring $\mA_7$.
Finally, when analyzing the random symmetric matrix $\mA_{11}$ (see \Cref{fig:lin_conv_fullyRandomA}), we discover that its associated operator coincidentally has spectral norm $0.9948$, which again coincides with the observed slow convergence properties.

\Cref{eqn:r2n2_linoperator} does not only allow to check for which problem instances a trained \R2N2 is applicable. 
Instead, it further lets us derive conditions on $\Vtheta$ such that the resulting \R2N2 is globally convergent on a linear problem class defined by one specific $\mA$ or a set of $\mA$. This should be explored in future work.